%% file: main.tex
\documentclass[]{fairmeta}
\title{\textsc{Decorum}: A Language-Based Approach For Style-Conditioned Synthesis of Indoor 3D Scenes}

\author[1,2,*]{Kelly O. Marshall}
\author[1]{Omid Poursaeed}
\author[1]{Sergiu Oprea}
\author[1]{Amit Kumar}
\author[3]{Anushrut Jignasu}
\author[2,\dagger]{Chinmay Hegde}
\author[1,\dagger]{Yilei Li}
\author[1,\dagger]{Rakesh Ranjan}

\affiliation[1]{Meta Reality Labs}
\affiliation[2]{New York University}
\affiliation[3]{Iowa State University}

\contribution[*]{Work done at Meta}
\contribution[\dagger]{Joint last author}

\abstract{3D indoor scene generation is an important problem for the design of digital and real-world environments. To automate this process, a scene generation model should be able to not only generate plausible scene layouts, but also take into consideration visual features and style preferences. Existing methods for this task exhibit very limited control over these attributes, only allowing text inputs in the form of simple object-level descriptions or pairwise spatial relationships. Our proposed method \textsc{Decorum} enables users to control the scene generation process with natural language by adopting language-based representations at each stage. This enables us to harness recent advancements in Large Language Models (LLMs) to model language-to-language mappings. In addition, we show that using a text-based representation allows us to select furniture for our scenes using a novel object retrieval method based on multimodal LLMs. Evaluations on the benchmark 3D-FRONT dataset show that our methods achieve improvements over existing work in text-conditioned scene synthesis and object retrieval.}

\date{\today}
\correspondence{Kelly Marshall at \email{km3888@nyu.edu}}

\metadata[Code]{Code will be released upon acceptance}
\usepackage{lineno}
\usepackage{amssymb}
\usepackage{bbm}
\usepackage{booktabs}
\usepackage{subcaption}
\usepackage{graphicx}
\usepackage{listings}
\usepackage{array}
\usepackage{graphicx}
\usepackage{adjustbox}
\usepackage{multirow}
\usepackage{makecell}
\usepackage{array}

\usepackage[most]{tcolorbox}
\tcbuselibrary{listingsutf8}

\newtcolorbox{promptbox}{
  colback=metabg,      % light background
  colframe=metafg,     % border color
  boxrule=0.5pt,
  arc=3pt,
  left=6pt,
  right=6pt,
  top=4pt,
  bottom=4pt,
  fontupper=\ttfamily\small\color{metafg},  % monospace + metafg color
  enhanced,
  breakable
}

\begin{document}

\maketitle

\input{sec/1_intro}
\input{sec/2_background}

\input{sec/3_textlayout}

\input{sec/4_decorate}

\input{sec/5_experiments}
\input{sec/6_conclusion}

\clearpage
\newpage
\bibliographystyle{assets/plainnat}
\bibliography{main}

\clearpage
\newpage
\beginappendix
\input{sec/X_suppl}
% \section{First appendix}

\end{document}

%% file: sec/1_intro.tex
\begin{figure*}[t!]
    \centering
    \includegraphics[width=0.99\linewidth,clip,trim={0.0in 0.5in 0.0in 0.0in}]{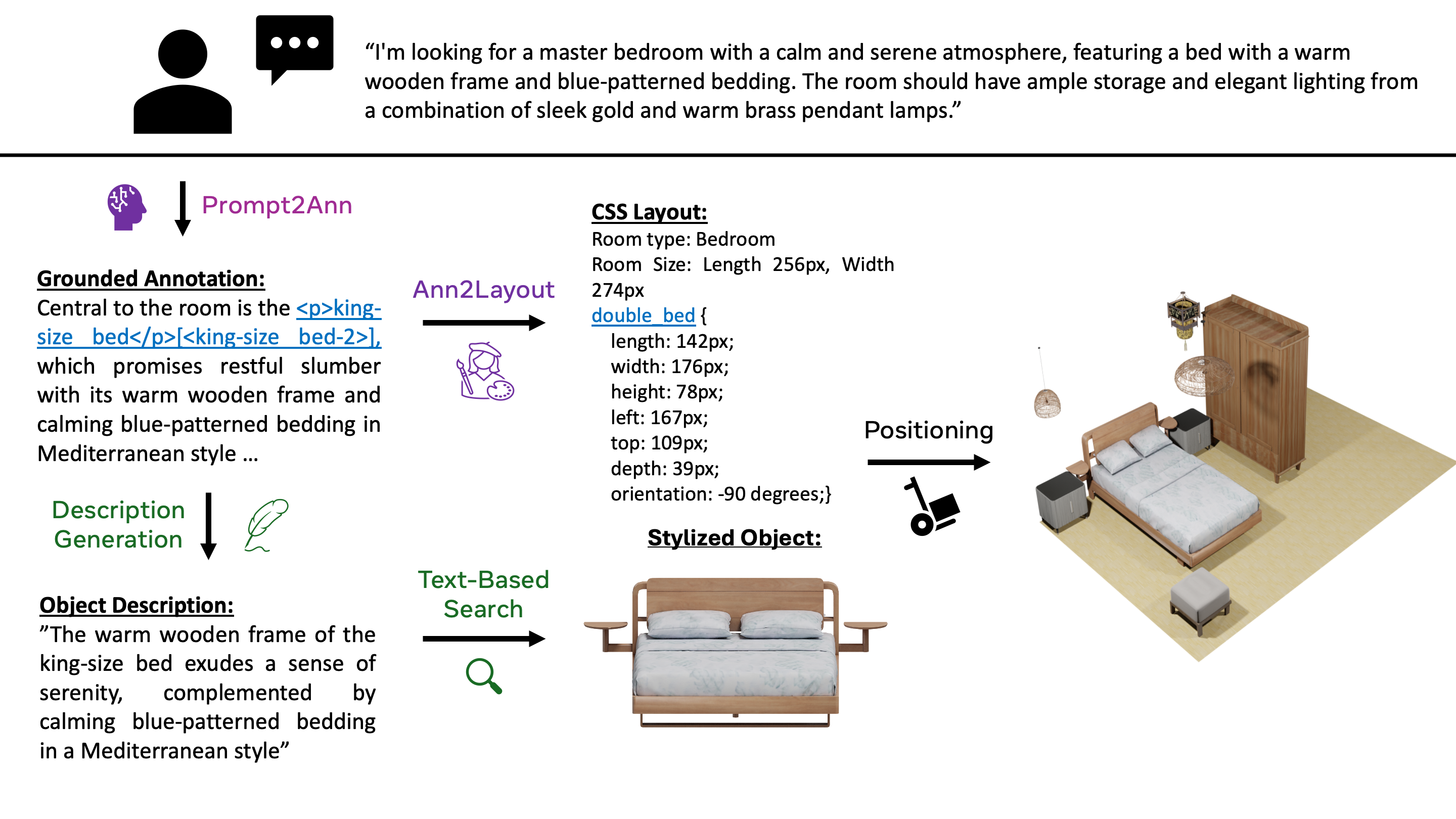}
    \caption{Illustration of the \textsc{Decorum} pipeline. We finetune LLaMA to obtain both Prompt2Annotation and Annotation2Layout models to convert from user prompts to CSS scene layouts in a two-stage process. Using this intermediate annotation representation decouples the generation of spatial and stylistic elements, allowing us to separately perform furniture selection using the tagged objects. Purple coloring indicates language model modules that we train using LoRA and green coloring shows our DecoRate furniture retrieval method which relies entirely on pretrained LLMs.}
    \label{fig:pipeline}
\end{figure*}

\section{Introduction}
\label{sec:intro}

The generation of 3D indoor scenes is a problem which has garnered considerable attention, with potential applications in video game development, automated interior design, and metaverse content creation \citep{merrell2011interactive,fisher2015activity,qi2018human,wang2018deep,ritchie2019fast,zhang2020deep,yang2021indoor,yang2021scene,hollein2023text2room,song2023roomdreamer,cohen2023set,lin2023componerf,LayoutGPT,patil2023advances}. In these applications, it is critical that the generation process produces scenes that are practical as well as visually pleasing. 

While existing works \cite{CLIPLayout,InstructScene} have taken into account visual coherence, this does not equate to producing visually pleasing scenes. This is because there does not exist a single objective standard for what constitutes an aesthetically appealing scene. Instead, it is a matter of individual preference. As a result, visual quality cannot be fully addressed by unconditional scene generation methods which do not incorporate user input.
% In these applications, it is critical that along with practical usability, generated scenes also possess aesthetic coherence and are visually pleasing. 

To better integrate user input, the scene generation process can be conditioned on natural language. This is a flexible and powerful interface which can allow users to describe their desired scenes in as much or as little detail as they desire. A description may include both style and spatial details, such as object and scene styles, as well as practical layout information. Existing text-conditioned methods lack the generality to handle both the style and structure of a scene in a text prompt, restricting user specifications to either brief object-level descriptions (e.g. color) \cite{CLIPLayout} or explicit pairwise spatial relationships \cite{InstructScene}. 

To fill this gap, we propose \textsc{Decorum}, a generative framework for synthesizing indoor 3D scenes from user prompts. \textsc{Decorum} decouples the generation of spatial and stylistic elements by first converting the user prompt into a dense annotation with object tags. Using this detailed annotation, \textsc{Decorum} separately designs a layout for the scene's objects and assigns each tagged object a textured mesh retrieved from an inventory of existing furniture. We provide an overview of our proposed model pipeline in Figure \ref{fig:pipeline}.

% Why text-based is good
Our method uses language-based representations for 3D scenes at each stage, allowing us to fully harness the power of Large Language Models (LLMs) towards scene generation. The scenes we output are in CSS format, following previous work \cite{LayoutGPT}  which shows that this structured representation facilitates spatial understanding in LLMs. Both the mapping from prompt to annotation and annotation to layout are learned using low rank adaptations \cite{LORA} of the open-source LLaMA 3.1 \cite{touvron2023llama}

Our use of language representations carries the additional benefit of allowing text-based search for furniture retrieval. Existing methods either generate only object category and size \cite{ATISS,LayoutGPT} or search over an inventory of textured meshes by using predicted object embeddings \cite{InstructScene,CLIPLayout}. Using the object tags in our generated scene annotation, we obtain detailed descriptions of each object in the scene, combining any descriptions of it included in the prompt with the information provided about the scene as a whole. This gives an informative object representation for performing text-based object retrieval. 

To make full use of this representation, we look at recent developments in Large Multimodal Models (LMMs) which ground LLMs' reasoning and world knowledge to visual information. Leveraging this link from textual to visual features, we devise a new scoring method to compute an object-text similarity metric which can be used for object-retrieval. We dub this retrieval method DecoRate, and show that its sensitivity to fine-grained image details leads to major improvements over CLIP-based object retrieval. To alleviate the computational burden of rating a potentially expansive inventory of objects with LMM inferences, we show the effectiveness of a coarse-to-fine pipeline which relies on CLIP to generate a small number of search candidates.

Integrating DecoRate into our \textsc{Decorum} pipeline results in a scene generation method which combines the spatial reasoning, natural language understanding, and multimodal capabilities of LLMs to comprehend and design scenes in line with varied user inputs. We evaluate \text{Decorum} on the benchmark 3D-FRONT dataset \cite{3DFRONT}, which consists of professionally arranged indoor scenes. Through quantitative and qualitative comparisons with existing works we show that our method has both the flexibility to represent a diverse range of scenes and the controllability to reflect a wide range of user inputs. We also separately verify the effectiveness of our DecoRate method for furniture retrieval.

% Existing approaches for generating stylistically cohesive scenes use transformer-based autoregressive methods and predict image embeddings to select relevant furniture objects.

% We propose making use of newly introduced paired scene-annotation data to import knowledge from the language domain. To do this, we follow existing work which shows that language models can effectively process 3D scene data when it is formatted as spatial layout code in the form of CSS. We use this as a spatial reasoning backbone for our generative model. In addition, we are able to use text descriptions to more effectively select objects to populate our generated scenes.

% "or have used graph based priors to model a scene's explicit spatial relationships"

% to convert user prompts first into a structured annotation which captures both explicit object categories as well as 

% By leveraging this semi-structured 

% fully harnessing pre-trained Large Language Models (LLMs). Spatial reasoning, natural language understanding, and multimodal [....]

\noindent \textbf{Paper Contributions.} The contributions of our work are summarized as follows:

\begin{itemize}
        \item  We propose \textsc{Decorum}, a method for synthesizing indoor 3D scenes from user descriptions by leveraging a semi-structured intermediate representation.
        % "the first method"
        \item We introduce DecoRate, a simple yet powerful text-based search method for object ranking and retrieval which achieves a 15x improvement in Top-1 retrieval accuracy compared to existing vector-similarity based methods.
        % \itemWe demonstrate the effectiveness of our methods using the benchmark 3D-FRONT dataset, outperforming existing scene generation methods in quantitative and qualitative comparisons. 
        \item We demonstrate that our pipeline for text-conditioned scene generation outperforms existing methods on the benchmark 3D-FRONT dataset. In addition, we introduce the Text Fidelity Ranking (TFR) metric to measure a scene generation model's adherence to text inputs.
\end{itemize}

%% file: sec/2_background.tex
\section{Background}
\label{sec:background}

\begin{figure*}[t!]
    \centering
    \includegraphics[width=\linewidth,clip,trim={0.0in 0.1in 0.0in 0.0in}]{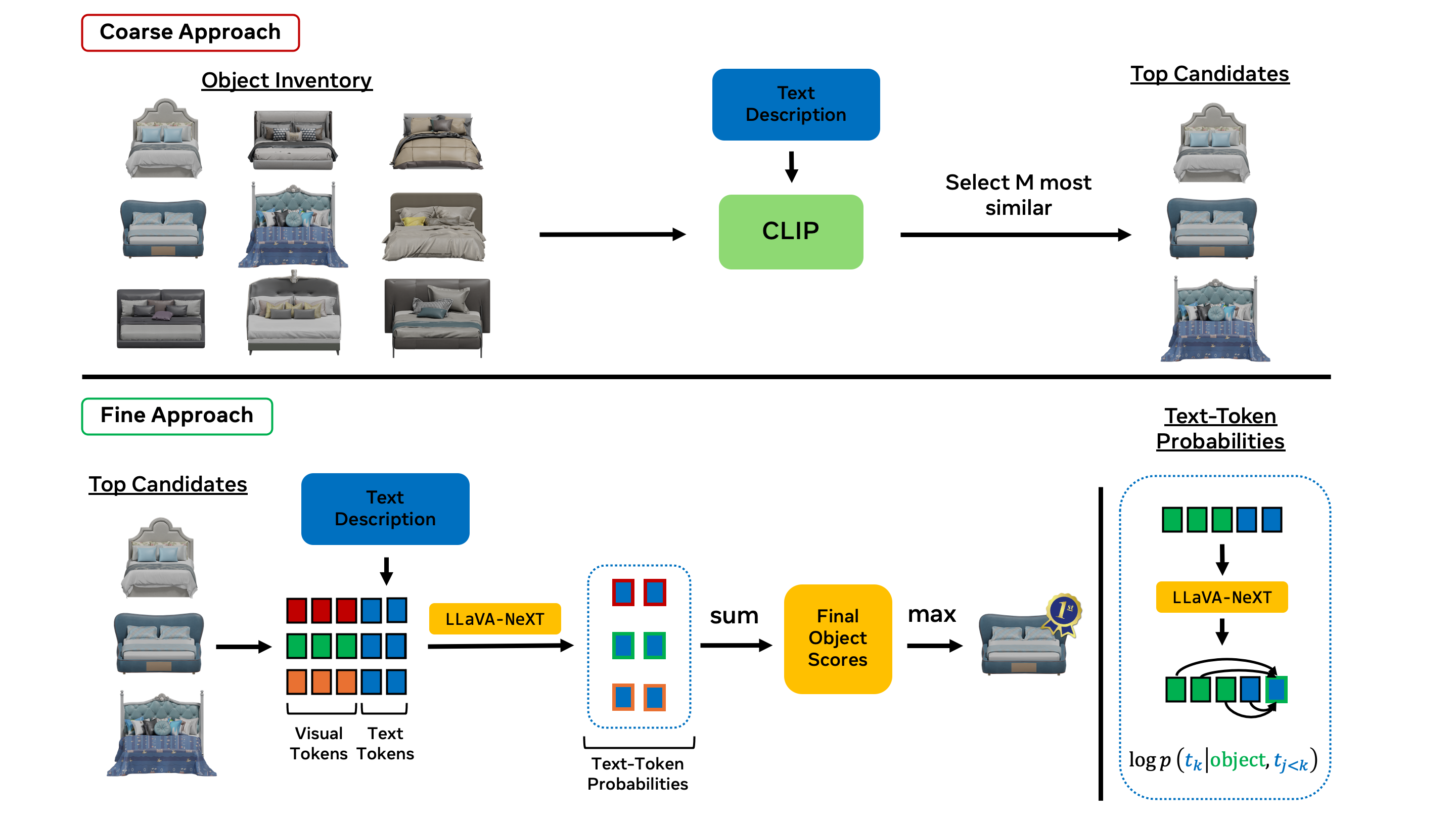}
    \caption{Illustration of the \textsc{DecoRate} coarse-to-fine rating system for text-based object retrieval. The top half shows our coarse CLIP-based candidate generation process. Below that is a visualization of our fine-grained rating system using LLaVA-NeXT. The dotted blue box contains our method for assigning probability scores to the text tokens (blue) conditioned on each input object’s visual tokens. These token probabilities are then summed together and added to the object prior probabilities (described in \ref{sec:prior}) to rate each object.}
    \label{fig:coarse_to_fine}
\end{figure*}

\noindent \textbf{Unconditional Scene Generation.}
Interest in the area of indoor scene generation has burgeoned following the release of the 3D-FRONT \cite{3DFRONT} dataset which contains several thousand professionally arranged scenes each consisting of a set of objects along with their associated translation and rotation values. This high-quality and large-scale dataset has enabled the development of more sophisticated generative modeling techniques applied to indoor scenes. While alternative datasets have been created using procedural content generation \cite{ProcTHOR,AI2THOR} data-driven generative models benefit from learning complex spatial relationships directly from human-designed layouts, leading to more realistic and semantically meaningful scene arrangements.

The most popular approach to generate these scenes has been a class of transformer-based methods \cite{ATISS} which autoregressively predict an object class and position given a partial scene. Separately, LayoutGPT \cite{LayoutGPT} has shown that by encoding the scene's spatial information into a CSS layout format, the unconditional generation problem can be solved by existing Large Language Models (LLMs). This approach requires no training, only contextual exemplars.

An enhanced version of the 3D-FRONT dataset, 3D-FUTURE \cite{3DFUTURE} was later released with a more extensive object catalog and textured meshes for each object. Subsequent works have thus focused on selecting objects based on their appearance and their contribution to the scene's aesthetic coherence\cite{CLIPLayout,InstructScene,diffuscene}.

\noindent \textbf{Text-Conditioned Scene Generation.}
Another line of work has sought to make indoor scene generation models responsive to text inputs.  CLIP-Layout \cite{CLIPLayout} builds on ATISS'\cite{ATISS} autoregressive structure by predicting a CLIP \cite{CLIP} embedding at each timestep, in addition to the object class, position, and rotation. This enables style-aware object retrieval using a nearest neighbor search of the object catalog's image embeddings. They are also able to perform text-based object replacement in which a new object can be selected during postprocessing based on the object description provided by an user. For scene-level text-conditioning, they can incorporate a single phrase (e.g "wooden", "colorful") at inference time by adding this phrase's vector embedding to each predicted CLIP embedding. This provides some user control over furniture selection but is extremely limited as 1) the text input does not influence the selection of object positions or categories and 2) a single CLIP embedding cannot translate a scene-level request into appropriate object selections.

Another such work is \cite{InstructScene} which focuses on producing scenes with well-defined spatial relationships between objects. The method works by extracting spatial relationships from 3D-FRONT scenes using hard-coded rules and then conditioning on these relation triplets. Using this dataset, they learn a two-phased diffusion model, using intermediate graph representation to map from natural language instructions to scenes. The resulting model is able to obey specified object relationships and object-level descriptions but does not follow scene-level style cues or general format natural language inputs.
% RelScene https://openreview.net/pdf?id=GIw7pmMPPX

More recently, the 3D-GRAND \cite{3DGRAND} dataset has paired 3D-FRONT scenes with detailed text annotations, providing a link between indoor scenes and object-level descriptions. Each annotation in the 3D-Grand dataset contains a tag and description for every object in its associated 3D-FRONT scene. Along with this, they contain descriptions of the scene's overall appearance, connecting object-level and scene-level attributes.

\noindent \textbf{Large Multimodal Models (LMMs).} Building on the recent success of Large Language models, an emergent line of work has aimed to integrate data from other modalities into LMM architectures \cite{BLIP2,flamingo,llava,llavanext}. In particular, vision-language models have been established as a method for encoding visual information into the representation space of pre-trained LLMs. By concatenating these visual tokens with a language input, LMMs allow for interaction between the multi-modal inputs within the attention mechanism of the LLM. As a result, the next-token predictions of LLMs can be further conditioned on an image input $x$ to get prediction probabilities $p(t_i|t_{k<i},x)$. Autoregressively sampling from this distribution allows for image-grounded text generation which can be used for image captioning, image-based question-answering, and other grounded reasoning tasks.

% One such LMM is LLaVa-NeXT\cite{llavanext} which uses a ViT 

%% file: sec/3_textlayout.tex
\begin{table*}[h]
    \centering
    \resizebox{0.75\textwidth}{!}{%
    \begin{tabular}{lcccc|c}
        \toprule
        & DecoRate & DecoRate$^{(100)}$ & DecoRate$^{(25)}$ & DecoRate$^{(10)}$ & CLIP \\
        \midrule
        Top-10 Accuracy & 47.3\% & 45.8\% & 28.1\% & 15.3\% & 15.3\% \\
        Top-5 Accuracy & 35.0\% & 34.5\% & 24.1\% & 15.3\% & 8.4\% \\
        Top-1 Accuracy & 16.3\% & 14.8\% & 11.3\% & 10.8\% & 0.5\% \\
        \midrule
        Runtime per Sample & 69.1s & 34.1s & 8.6s & 3.5s & 0.01s \\
        \bottomrule
    \end{tabular}
    }
    \caption{Top-K Accuracy for retrieving ground truth furniture using different text-based furniture retrieval methods. DecoRate$^{(M)}$ denotes our DecoRate search method applied to the top $M$ results returned by CLIP-based retrieval.}
    \vspace{-0.2cm}
    \label{tab:TopK}
\end{table*}

\section{Decorum}
\label{sec:methods}

% Taking advantage of densely grounded captions as a semi-structured middleground between natural language inputs and highly structured code outputs. 

In this section we outline key parts of \textsc{Decorum}, our approach for text-conditioned scene generation which relies on an intermediate representation in the form of densely grounded scene annotations. First, in \ref{sec:layout_gen} we introduce the layout generation model which produces a structured representation of a scene's object layout based on a detailed scene annotation. This relies upon the ability to convert from user prompts to this scene annotation format which is handled by the annotation generation model described in \ref{sec:ann_gen}. This generation pipeline is displayed in Fig. \ref{fig:pipeline}.

We leave the details of our furniture selection method DecoRate, which uses scene annotations to furnish the generated layouts with textured meshes, to Sec \ref{sec:decorate}. 

\subsection{Problem Statement}

Following the notation established in \cite{ATISS}, we denote a 3D scene $\mathcal{X}_i$ as a tuple $\mathcal{X}_i = \left( \mathcal{O}_i, \textbf{F}^i \right)$ containing a set of objects $\mathcal{O}_i = \{o_j^i \}_{j=1}^M$ and a floor layout $\textbf{F}^i$. Each object within the scene contains attributes for category, size, orientation and location. In our setting, we also associate each object with a textured mesh $\tau$ belonging to an inventory $\mathcal{I}$. 

Given a short (2-3 sentences) user description partially or completely describing a room, along with its 2D floorplan $\textbf{F}$, our method determines a set of objects $\mathcal{O}$ to drag and drop onto the given floorplan - creating a scene that matches the structure and style specified in the input.

% This scene should consist of a set of objects, each paired with a mesh in the 3D-FUTURE catalog along with a layout which describes the objects' spatial positioning.

\subsection{Layout Generation Model}\label{sec:layout_gen}

In order to convert between natural language and 3D scenes, we propose training a model to predict 3D-FRONT \cite{3DFRONT} layouts based on their corresponding 3D-GRAND \cite{3DGRAND} captions, which we refer to as Ann2Layout. This conversion is eased by the highly structured format of 3D-GRAND, which allows the model to retrieve the list of objects present in the layout.

Following LayoutGPT \cite{LayoutGPT}, we represent the layout's spatial information in CSS format and use a language model which has gained extensive familiarity with CSS through its training corpus. However, unlike LayoutGPT which relies only on in-context learning to avoid learning model parameters, we train a low-rank adapter \cite{LORA} on top of the open-source LLaMA 3.1 \cite{touvron2023llama} model. This enables the model to benefit from the paired data at our disposal ensuring it learns to generate layouts which reflect the annotation input while also modeling spatial relationships between objects. Furthermore, with highly structured CSS code as the output of the model, we are able to generate all the attributes of each object $o_j^i$ except the textured mesh.

\subsection{Annotation Generation Model}\label{sec:ann_gen}

While a 3D-GRAND style annotation is a significant step in the direction of natural language from the 3D domain, it still leaves much to be desired in terms of friendliness to users. Most design specifications do not come formatted as detailed grounded annotations with each object specified and tagged. Instead, a user may wish to describe the room in more natural terms, perhaps describing stylistic elements which are important to them or specifying some subset of the desired objects in the room.

To make this possible, we create a prompt-to-annotation model, called Prompt2Ann, to act as the user-facing portion of our scene generation pipeline. This model produces a 3D-GRAND-style dense scene annotation to match the specifications of a given user prompt. We once again finetune LLaMA 3.1 for this task using LoRA.

To train our model, we require a paired dataset containing 3D-GRAND annotations along with user prompts which could plausibly be used to generate them. As we already possess the 3D-GRAND annotations, we simply generate matching user prompts by asking the base LLaMA 3.1 model to summarize each 3D-Grand annotation and pose its summary as a user prompt (we include details of our prompting strategy in the Appendix). After creating this paired dataset, we then finetune our adapter to map from prompts to annotations. 

With these prompt-conditioned annotations, we now have a suitable input for the Ann2Layout model described in Sec. \ref{sec:layout_gen}. In addition, this representation retains all of the aesthetic information necessary to furnish the scene layout using our furniture retrieval method DecoRate, which we now describe in Sec. \ref{sec:decorate}.

%% file: sec/4_decorate.tex
\section{DecoRate}
\label{sec:decorate}

This section contains a description of the novel LLM-based object furniture retrieval method DecoRate. DecoRate takes as input a densely grounded scene annotation, which describes the room in detail and includes explicit tags for each room object. By translating this annotation into a sequence of object descriptions, we can use text-based search to retrieve textured meshes from our inventory $\mathcal{I}$ for all of the scene's furniture. Because our layout generation model is also conditioned on this scene representation, we can then use these meshes to furnish our generated layout, satisfying both spatial and visual specifications.

We first detail our method for converting scene annotations into object-level descriptions in Sec.\ref{sec:descriptions}. Using these descriptions, we derive a novel object rating system to allow high-quality text-based search using LLMs, derived in Sec.\ref{sec:llm_stylize}. Based on this derivation, we also include a frequency-based object prior, which we discuss in Sec.\ref{sec:prior}. Finally, in Sec.\ref{sec:coarse_to_fine} we address the runtime of our proposed method and show how we can use a two-stage coarse-to-fine method to trade off between retrieval quality and runtime. This coarse-to-fine strategy is visualized in Fig. \ref{fig:coarse_to_fine}.

\subsection{Description Generation}
\label{sec:descriptions}

In order to convert scene annotation inputs into a format that is compatible with our text-based object retrieval method (described in Sec.\ref{sec:llm_stylize}) we must first extract short text descriptions for each object in the description. One approach for getting this description for each tagged object would be to simply select all of the relevant text surrounding the object tag. However, this would not properly capture scene-level attributes which influence the object selection process. For instance, a user might not include a description of the type of wardrobe they would like in their room, but instead state that they desire a room with a modern aesthetic.

To ensure that both scene-level and object-level information is propagated into furniture styles, we use a pre-trained LLaMA 3.1 model \cite{touvron2023llama} to generate each object description based on the scene annotation. We provide a prompt instructing the model to output a description following formatting guidelines for each tagged object (prompt details are left to the Appendix) followed by both the user prompt and scene annotation. We find that the pre-trained model is capable of consistently generating descriptions for each object in the scene and that these descriptions succinctly capture relevant visual features based on both scene-level and object-level style information.

\subsection{LMM Object Rating}
\label{sec:llm_stylize}

% To enable style-driven scene generation, it is imperative that the outputted scene not only contain the correct object categories in a sensible layout, but that each object reflect the style and aesthetic specified as input. Accomplishing this without individually generating each scene asset requires a procedure for assigning each object a textured mesh from a predetermined inventory $\mathcal{I}$. The chosen mesh should conform to the scene-level aesthetic as well as any specific object-level attributes present in the input text. 

% To accomplish this, we adopt a two-phase approach. First, we use our densely annotated scene annotation to generate a text description of each object in the scene. Then, for each item in the scene we leverage a pretrained multimodal LLM to rate each element in $\mathcal{I}$ based on how well its appearance matches the generated description.
In this section we describe our novel method for assigning probability-based scores to objects within an inventory based on their agreement with a given text prompt. To do this, we begin with a straightforward maximum likelihood objective and derive an equivalent formulation which can be well-approximated by a multimodal LLM.

Formally, given a text description $t$ we would like to find the object which best matches our description from an existing set $\mathcal{I}=\{\tau\}_{i=1}^N$. As each object may fulfill the description of the text prompt to varying degrees, we frame this as a likelihood maximization, equivalent to finding:

$$\tau^*=\max_{\tau_i \in \mathcal{I}} p(\tau_i|t)$$

% Computing $p(\tau_i|t)$ exactly is a bit tricky. One way would be to use CLIP embeddings and define:

% $$\hat{p}(\tau_i|t) = \frac{\left<E(\tau_i),E(t)\right>}{\sum_{j=1}^N \left< E(o_j),E(t)\right>}$$

% Then the denominator can be ignored and the optimal object can be found with just $N$ dot product computations. While this is fast, its reliance on CLIP makes it relatively insensitive to details and fine-grained semantics which may be present in $t$. 

In this form, $p(\tau_i|t)$ is intractable to compute as it requires a likelihood-based generative model over the discrete set $\mathcal{I}$. However, by applying Bayes' rule and, we can obtain an expression which instead depends on the conditional probability $p(t|\tau_i)$. 

% $$\tau^*=\max_{\tau_i \in \mathcal{I}} \frac{p(t|\tau_i)p(\tau_i)}{p(t)}$$

% As $p(t)$ is a constant, we get the simplified expression:

$$\tau^*=\max_{\tau_i \in \mathcal{I}} p(t|\tau_i) p(\tau_i)$$

This is a simpler form as it is straightforward to calculate $p(t|\tau_i)$ with a multimodal LLM as the product of next-token probabilites and the prior $p(\tau_i)$ (we provide more detail about $p(\tau_i)$ in Sec.\ref{sec:prior}). By converting $t$ into tokens $x_1,\cdots,x_T$ we can determine $\tau^*$ as follows:

$$\tau^* = \max_{\tau_i \in \mathcal{I}} p(\tau_i)\prod_{j=1}^N p_{LLM}(x_j|\tau_i,x_{k<j})$$

This yields an object scoring objective which is both principled and tractable, allowing us to define a novel text-based retrieval method which leverages the capacities of LLMs. 

$$\tau^* = \max_{\tau_i \in \mathcal{I}} \log(p(\tau_i)) + \sum_{j=1}^N \log (p_{LLM}(x_j|\tau_i,x_{k<j})) $$

In practice, we find it improves performance to introduce a weighting coefficient $\lambda_p$ to control the trade off between the prior and conditional terms. We provide an ablation of the $\lambda_p$ hyperparameter in Tab.\ref{tab:prior_ablation}.

% Table \ref{tab:prior_ablation}

% $$\tau^* = \max_{\tau_i \in \mathcal{I}} \underbrace{\lambda_p \log(p(\tau_i))}_{\text{Prior}} + \underbrace{\sum_{j=1}^N \log (p_{LLM}(x_j|\tau_i,x_{k<j}))}_{\text{Conditional}} $$

% $$\tau^* = \max_{\tau_i \in \mathcal{I}} \lambda_p \log(p(\tau_i)) + \sum_{j=1}^N \log (p_{LLM}(x_j|\tau_i,x_{k<j}))$$

\subsection{Object Prior}
\label{sec:prior}

To estimate the unconditional prior over $p(\tau)$ for $\tau\in\mathcal{I}$, one option is to assume a uniform prior $p(\tau)=\frac{1}{|\mathcal{I}|}$, equivalent to setting $\lambda_p=0$. We demonstrate the effectiveness of this simple data-free approach in Tab.\ref{tab:prior_ablation}.

Alternatively, we can use the object frequencies in the training set to estimate a categorical distribution. To avoid overfitting, we begin with a uniform Dirichlet prior and update $p(\tau)$ in a Bayesian manner \cite{drichlet}. This results in an estimate of $p(\tau)$ which reflects the training dataset while also ensuring that its support contains all possible objects.

% We select the Dirichlet distribution as our conjugate prior andStarting with the uniform Dirichlet distribution as a conjugate prior, we update the distribution with each observed sample . 

\subsection{Coarse-To-Fine}
\label{sec:coarse_to_fine}

While DecoRate gives a better estimation of the optimal object for retrieval due to its more fine-grained understanding of language structure, it comes with a significant computational cost as it requires $N$ forward passes through whichever LLM we use. This can be somewhat alleviated by implementation optimizations which eliminate redundant data transfers to GPU (see Appendix for details) without sacrificing performance. 

However, for more significant speedups at the expense of some retrieval accuracy, we propose using CLIP to pre-filter $\mathcal{I}$, thereby reducing the number of LLM inference calls for each object retrieval to a fixed number. This coarse-to-fine approach allows us to take advantage of the low runtime of CLIP-based search while retaining the fine-grained image understanding of LLM-based rating. The resulting two-phase search method is shown in Fig. \ref{fig:coarse_to_fine}

% However, this runtime can be improved through inference optimizations as well as filtering strategies to reduce the number of candidate objects to a small fraction of $N$.

\begin{figure*}[htbp]
    \centering
    \includegraphics[width=\textwidth]{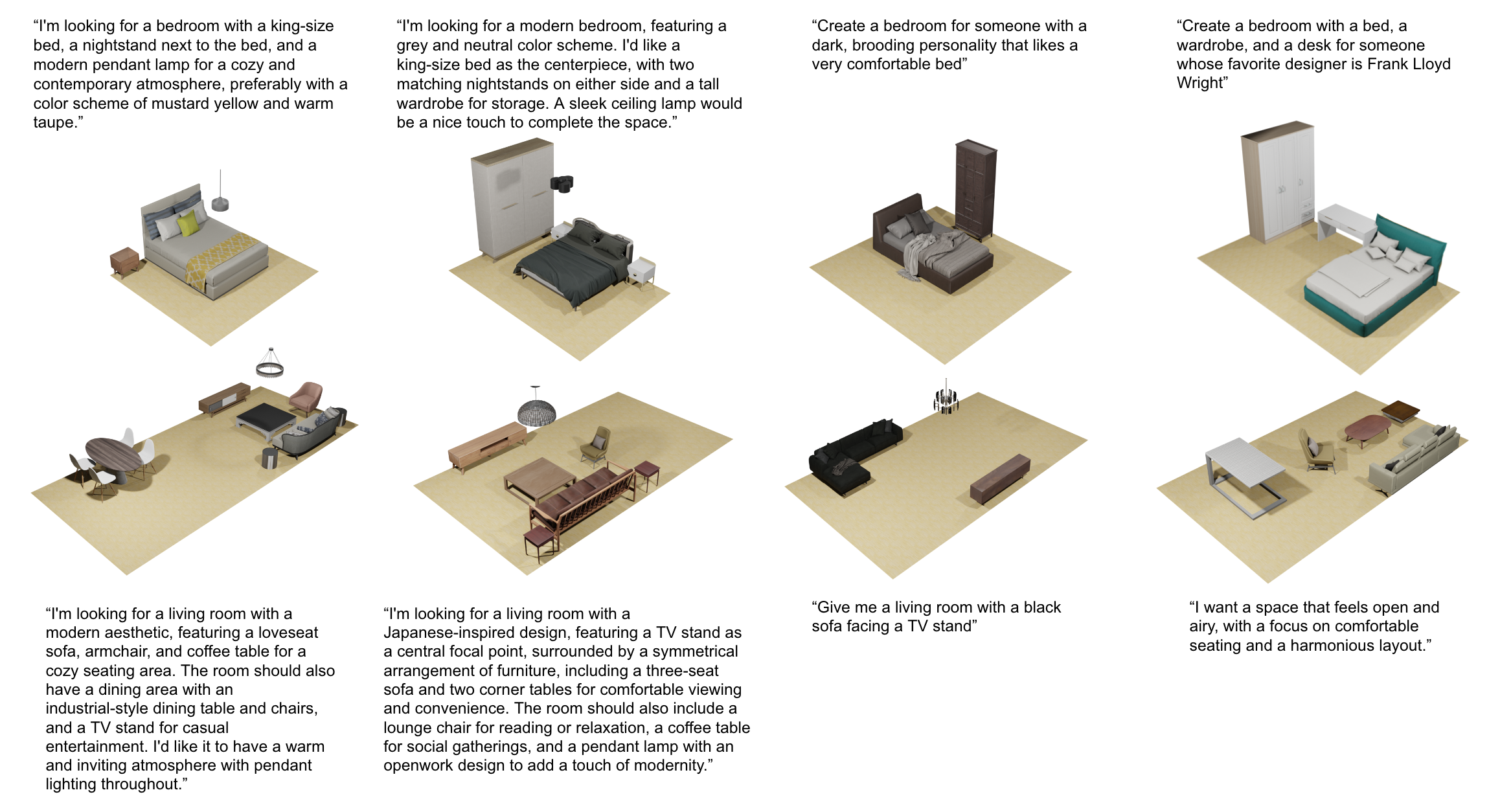}
    % \vspace{0.5cm} % Adjust the space between the images as needed
    % \includegraphics[width=\textwidth]{Figures/lr_qualitative.png}
    \caption{Examples of prompt-conditioned scene generation for bedrooms (top) and living rooms (bottom) with \textsc{Decorum}. We condition on prompts taken from the test set generated by LLaMa (left) and out-of-distribution prompts in a different format (right). We show that our method can accurately generate 3D indoor scenes based on user prompts that specify both spatial and stylistic attributes.}
    \label{fig:qualitative}
\end{figure*}

%% file: sec/5_experiments.tex
\begin{table*}[h]
    \centering
    \resizebox{0.75\textwidth}{!}{%
    \begin{tabular}{lccccc|c}
        \toprule
        % & \multicolumn{6}{c}{Textual Fidelity Rate (TFR) Scores} \\
        % \cmidrule(lr){2-7}
        Method & TFR@0.5 & TFR@0.8 & TFR@0.9 & TFR@0.95 & TFR@1.0 & Mean TFR \\
        \midrule
        \multicolumn{7}{c}{\textbf{Bedrooms}} \\
        \midrule
        Ground Truth & 0.98 & 0.82 & 0.8 & 0.64 & 0.42 & 91.8\% \\
        \textbf{Decorum (ours)} & 0.72 & 0.33 & 0.28 & 0.10 & 0.08 & 66.6\% \\
        ClipLayout & 0.5 & 0.14 & 0.08 & 0.06 & 0.0 & 50.0\% \\
        % Placeholder values for other methods
        % Add more methods here
        \midrule
        \multicolumn{7}{c}{\textbf{Living Rooms}} \\
        \midrule
        Ground Truth & 0.89 & 0.60 & 0.43 & 0.20 & 0.06 & 78\% \\
        \textbf{Decorum (ours)} & 0.84 & 0.50 & 0.34 & 0.12 & 0.02 & 72.2\% \\
        ClipLayout & 0.5 & 0.10 & 0.06 & 0.0 & 0.0 & 50.0\% \\
        % Placeholder values for other methods
        % Add more methods here
        \bottomrule
    \end{tabular}
    }
    \caption{Comparison of Textual Fidelity Rate (TFR) scores for Bedrooms and Living Rooms.}
    \vspace{-0.2cm}
    \label{tab:TFR}
\end{table*}

\begin{table*}[h]
    \centering
    \resizebox{0.75\textwidth}{!}{%
    \begin{tabular}{lccc|ccc}
        \toprule
        & \multicolumn{3}{c|}{Bedrooms} & \multicolumn{3}{c}{Living Rooms} \\
        \cmidrule(lr){2-4} \cmidrule(lr){5-7}
        Method & FID $\downarrow$ & CLIP-FID $\downarrow$ & KID $\downarrow$ & FID $\downarrow$ & CLIP-FID $\downarrow$ & KID $\downarrow$ \\
        \midrule
        \textbf{Decorum (ours)} & \textbf{18.0} & \textbf{1.7} & \textbf{0.005} & \textbf{90.9} & 7.39 & 0.053 \\
        InstructScene & 21.0 & 2.4 & 0.007 & 92.6 & \textbf{6.85} & 0.025 \\
        CLIPLayout & 22.5 & 2.5 & 0.006 & 92.6 & 7.08 & \textbf{0.017} \\
        ATISS & 38.7 & 5.0 & 0.008 & 100.9 & 10.4 & 0.025 \\
        LayoutGPT & 44.3 & 7.0 & 0.025 & 99.4 & 10.8 & 0.045 \\
        LayoutGPT w/ prompts & 42.7 & 6.9 & 0.025 & 94.3 & 10.1 & 0.044 \\
        \bottomrule
    \end{tabular}
    }
    \caption{Comparison of FID, CLIP-FID, and KID scores for Bedrooms and Living Rooms.}
    \vspace{-0.2cm}
    \label{tab:FID}
\end{table*}
\section{Experiments}
\label{sec:experiments}

\subsection{Datasets}

We use the 3D-FRONT \cite{3DFRONT} dataset for each of our experiments. 3D-FRONT contains 18,000 rooms with furniture arranged by human designers to have plausible and practical layout configurations. The 3D-FUTURE \cite{3DFUTURE} dataset provides furniture objects for each 3D-FRONT room in the form of 10,000 textured meshes.

To obtain ground-truth densely-grounded annotations for 3D-FRONT scenes, we look to the 3D-GRAND \cite{3DGRAND} dataset which provides several types of scene annotations, including grounded scene description, grounded question-answering, and grounded object reference. For our purpose, we consider only the set of grounded scene descriptions, of which there are 65,000 in total.

Like LayoutGPT, our layout generation module can only process rectangular shaped scenes. As a result, we follow LayoutGPT's data filtering strategy for 3D-FRONT resulting in a train/val/test split of 3397/453/423 for bedroom scenes and 690/98/53 for living room scenes.

\subsection{Quantitative Evaluations}

\noindent \textbf{Style Retrieval Metrics.} To assess the effectiveness of our DecoRate method for furniture selection compared to CLIP-based retrieval we measure each method's accuracy in retrieving ground-truth objects from 3D-FRONT scenes based on 3D-GRAND captions. To accomplish this, we use our description generation method to acquire object-level descriptions of each object. We then use these descriptions to rank all object meshes in the 3D-FUTURE catalog and measure accuracy based on whether the Top-K ranked scenes include the ground truth textured mesh.

Table \ref{tab:TopK} reports Top-K accuracy for CLIP and DecoRate models applied to varying numbers of CLIP-produced candidates and reports their respective runtimes, using a total of $200$ ground truth objects. For each DecoRate model we set $\lambda_p=0.1$. Our results show that DecoRate is far more effective than CLIP based methods for style retrieval. In particular, its improved sensitivity to fine-grained details results in a much higher success rate for producing an exact match based on attributes specified through text. 

While CLIP performs poorly at distinguishing between objects which have relatively high similarity to the text prompt, it is effective as a coarse filter for ruling out irrelevant candidates. When reducing the number of LLM ratings to just 10, CLIP is able to retain enough relevant objects that DecoRate achieves a Top-1 accuracy 20x higher than CLIP alone while simultaneously seeing a 20x reduction in runtime compared to DecoRate by itself.

% To validate our choice of the hyperparameter $\lambda_p$, we include an ablation of DecoRate's performance using different prior weightings. The results in Table \ref{tab:prior_ablation} show that a value of $\lambda_p=0.1$ leads to optimal performance.

\noindent \textbf{Prompt-Conditioned Scene Generation.} We put our full Text-to-Scene pipeline to the test by generating and furnishing 3D scenes for each of the LLaMA-generated user prompts derived from the 3D-GRAND test set. Along with each prompt, the model is conditioned on a room layout specified in length and width dimensions.

Following previous works \cite{ATISS,LayoutGPT,CLIPLayout,InstructScene} we compute image-space metrics by rendering eight views of each generated scene. We compare these renderings to those of the ground-truth test set by measuring Fréchet Inception Distance (FID), CLIP-FID, and Kernel Inception Distance (KID) \cite{cleanfid}. We present these results in Table \ref{tab:FID}. 

When computing these metrics for existing works which do not use user input, we condition only on the room layout. This includes CLIPLayout which does not condition on text for their metric computations. For InstructScene, we also condition on the spatial relationship triplets and object-level descriptions provided in the InstructScene dataset. Our LayoutGPT implementation uses the authors' official repository with GPT API calls replaced by LLaMA 3.1 inference. We additionally report results for LayoutGPT with our scene summaries added to each prompt. 

Our quantitative evaluations (shown in Tab. \ref{tab:FID}) show that \textsc{Decorum} is competitive with existing methods in capturing the overall scene distribution. In particular, our method achieves the best performance in all metrics for bedrooms and outperforms all other baselines on FID for living rooms.

% GT: 0.265869140625
% Random: 0.258544921875
% Ours: 0.259033203125

\noindent \textbf{Text Fidelity Rate.} While inception-distance based metrics verify that our model is able to approximate the overall scene distribution of 3D-FRONT, we require a way to measure how well the generated scenes agree with their corresponding text prompts.
One such metric is CLIP-Score \cite{CLIPScore}. 

However, given that we've observed CLIP's limited sensitivity to fine-grained object characteristics when retrieving individual objects, it is unreasonable to expect that it could pick up on these same details when evaluating at a scene level. Instead, to get a more detailed evaluation we introduce an additional metric for text-conditioned scene synthesis based on multimodal LLMs (we include additional CLIP-Score evaluation results in the Appendix).

Textual Fidelity Rate (TFR) measures how clearly a model's output reflects its input text prompt based on the likelihood that an LLM could autoregressively reconstruct the text prompt from a rendering of the scene. This results in a holistic metric which quantifies how much textual information is retained through the scene generation process. 

To compute TFR for a scene-text pair, we first score the likelihood of this positive pair as the probability of generating the prompt given the scene rendering. We then compare this score to the likelihood scores for a randomly sampled batch of negative samples $\mathcal{N}$ generated by the model. Finally, we calculate TFR as the proportion of negative pairs with lower probabilities than the positive pair.

$$ \text{TFR}(\mathcal{S},t) = \frac{1}{|\mathcal{N}|} \sum_{\mathcal{S}^-\in \mathcal{N}} \mathbbm{1}_{p(t|\mathcal{S})<p(t|\mathcal{S}^+)} $$

To ensure diversity in views we again use eight-view scene renders. When computing the score for the positive pairs we take the median score over all eight-views to reduce noise from obstructed views. We further define the TFR@$k$ metric as the proportion of samples with a TFR score greater than or equal to $k$.

Table \ref{tab:TFR} provides TFR metrics on our test set of user prompts derived from 3D-GRAND. For each method, we compute TFR for $50$ scenes, with a negative sampling batch size of $50$ for each positive pair.

For comparison, we evaluate \textsc{Decorum} alongside CLIPLayout as well as ground-truth 3D-FRONT scenes corresponding to text prompts. We omit the InstructScene model in this comparison because it was not designed to accept general-form natural language input. When evaluating CLIPLayout we use their method for text-conditioned scene synthesis which conditions on a CLIP embedding of the text prompt. We find that CLIPLayout's performance is on par with random chance, indicating that it is unable to properly represent the scene's stylistic contents with just a single CLIP embedding. By contrast, \textsc{Decorum} demonstrates the ability to accurately represent textual inputs, leading to distinguishable text-based scene generation.

\subsection{Qualitative Results}

\noindent \textbf{Prompt-Conditioned Scene Generation.} 
In this section we present visual results for prompt-conditioned scene synthesis with our model \textsc{Decorum}. In this setting, the model is provided with a room layout specified by width and height dimensions as well as a short user prompt detailing desired attributes for the room's contents. 

Figure \ref{fig:qualitative} displays example prompts taken from the test set as well as out-of-distribution text inputs paired with \textsc{Decorum}'s scene outputs for these prompts. We also additional visual examples in Appendix Sec \ref{sec:qual_examples}. Notably, \textsc{Decorum} is able to translate the natural language specifications in the input prompts into appropriate furniture choices while maintaining plausible layouts due to its capacity for spatial representation. In addition, our use of pretrained LLMs allows for generalization to unseen prompt formats and use of world knowledge (e.g Frank Lloyd Wright is associated with the Modernist style).

% \begin{table}[h]

% \centering

% \begin{tabular}{|l|c|c|c|}

% \hline

% \textbf{Method} & \textbf{FID}  \\

% \hline

% Decorum(ours) & 18 \\

% CLIPLayout & 22  \\

% \hline

% \end{tabular}

% \caption{Ablation study results showing the impact of different features on model performance.}
% \label{tab:FID}

% \end{table}

%% file: sec/6_conclusion.tex
\section{Conclusion}
\label{sec:conclusion}

In this work, we introduce \textsc{Decorum}, a fully language-based pipeline for generating indoor 3D scenes based on text input. We show that by using densely-grounded annotations as an intermediate representation we are able to account for both stylistic and spatial specifications from users. To the best of our knowledge, \textsc{Decorum} is the first scene generation model to condition on language descriptions of both scene-level and object-level style attributes.

As part of our \textsc{Decorum} pipeline, we introduce DecoRate, a novel method for text-based object retrieval which we derive from a maximum likelihood objective. This method allows us to better capture object details in the retrieval process, leading to orders of magnitude improvement in retrieval accuracy along with greater visual quality.

We hope that our model will enable a new level of user interactivity and visual creativity in the world of 3D environments. In future work, we plan to extend our layout formulation to represent non-regular scenes and apply our innovations in text-based item retrieval to other domains.

%% file: sec/X_suppl.tex
\setcounter{page}{1}
% \maketitlesupplementary

\section{Implementation Details}

\subsection{LLM Prompting}
In this section we give more detail on our usage of pretrained LLMs for various components in our model pipeline. Other than the multimodal LLaVa-Next model that we use for DecoRate, all LLMs used are derived from the pretrained instruction-tuned LLaMa-3.1 8B parameter model. Below, we list the prompts which we prepend to model inputs using LLaMa's chat template.

\noindent \textbf{Summarization:} We use the following prompt with a pretrained model to summarize 3D-GRAND captions into user-style prompts.
\begin{promptbox}
Summarize the following grounded annotation into a shorter description that a user would use to describe what type of room they would like for their home:
\end{promptbox}

\noindent \textbf{Description:} The following prompt is used for generating descriptions for each tagged object in a densely grounded scene annotation.
\begin{promptbox}
I'm going to give you a detailed scene annotation with object tags. For each tagged object give me a detailed object description so that all the objects match the overall theme of the room and any description details in the annotation. Put each object description on its own line (if there are several objects which are the same just repeat the same description) in the format [object tag]: [description]. Don't spend more than three sentences on a single object. Do not include the explicit object tag (e.g <wardrobe-0>) in your description just use natural language. Output the objects in the same order that they are listed in the scene annotation. Detailed annotation:
\end{promptbox}

\noindent \textbf{Layout Generation:} 
The following prompt is inputted to our Ann2Layout module during both training and inference.
\begin{promptbox}
Make a room layout in CSS format that matches the following annotation:
\end{promptbox}

\noindent \textbf{Annotation Generation:} 
The following prompt is inputted to our Prompt2Ann module during both training and inference.
\begin{promptbox}
Generate a detailed room annotation with object tags from the following short user prompt:
\end{promptbox}

\noindent \textbf{DecoRate scoring:} 
When using the DecoRate model for scoring object-text agreements, we prompt the model for object description by including the following prompt with each input image before supplying the text description:
\begin{promptbox}
What is shown in this image?
\end{promptbox}

\begin{table*}[h]
\centering
\begin{tabular}{|l|c|c|c|c|c|c|}
\hline
 & $\lambda_p=0.0$ & $\lambda_p=0.01$ & $\lambda_p=0.1$ & $\lambda_p=0.5$ & $\lambda_p=1.0$ \\
\hline
Top-10 Accuracy & 27.6\% & 30.5\% & 47.3\% & 39.9\% & 33.0\%  \\
Top-5 Accuracy & 18.2\% & 22.2\% & 35.0\% & 26.1\% & 21.7\%  \\
Top-1 Accuracy & 5.4\% & 6.9\%  & 16.3\% & 10.3\% & 9.9\% \\
\hline
\end{tabular}
\caption{Comparison of different values of $\lambda_p$ for controlling the strength of the prior distribution term in DecoRate's scoring function. }%$\lambda_p=\infty$ denotes that we are scoring based only on the prior without taking into account the LLM output}
\label{tab:prior_ablation}
\end{table*}

\subsection{Model Training}

In this section we give implementation details for the finetuning of our Prompt2Ann and Ann2Layout models. For all experiments, we use a LORA alpha of 8, a learning rate of $0.0003$ and an L2 regularization of $0.001$. All finetuning was performed on a single 40GB NVIDIA A100. In order to achieve a higher effective batch size using limited GPU memory we use 8 gradient accumulation steps per update, each with a batch size of 1.

For our Prompt2Ann model, we use a LORA rank of 16 for all training runs. Training our model to convergence under this setting takes around 48 hours.

For our Ann2Layout model we use a LORA rank of 16 when training on bedroom scenes and for living room scenes we use a LORA rank of 64. Both types of scenes require around 4 hours of training to converge.

\subsection{DecoRate Runtime Optimization}

In this section we give more detail on how we optimize DecoRate's runtime to avoid redundant computations and GPU transfers. Given a sequence of text prompts $t_1,...,t_n$ where each $t_i$ is associated with $k$ candidate objects $\tau_1^{(i)},...,\tau_k^{(i)}$, we wish to compute $p(t_i|\tau_j),\forall i,j$ while minimizing file I/O.

To do this, we process a single object category at a time, precomputing all relevant features and storing them on GPU. For each category, we first find the set of all objects which are a candidate for at least one text description within that category. We then load the image files for these objects, transfer the images to GPU memory, and use LLaVa-Next's image encoder to compute their visual features. We then iterate over all of the relevant text descriptions, gathering any relevant precomputed features to compute the necessary likelihood scores. Once we are finished with this, we delete all of the visual features for the category from GPU memory.

This method improves DecoRate runtime by $\sim$20\% by ensuring that each object only has to be loaded, transferred to GPU, and encoded just once. Furthermore, these efficiency gains become more pronounced when using a larger set of inputs, improving the scalability of our method.

\section{Additional Results}

\subsection{$\lambda_p$ Ablation}
In Table \ref{tab:prior_ablation} we include an ablation of the $\lambda_p$ hyperparameter introduced in Sec. \ref{sec:llm_stylize}. We find that setting $\lambda_p=0.01$ gives the best results and as a result we use this choice of hyperparameter for our other experiments.

\subsection{Qualititative Comparison of Object Retrieval using DecoRate vs. CLIP}

In Sec. \ref{sec:experiments} we showed quantitative results to support the effectiveness of DecoRate for object retrieval compared to CLIP-based search. As further evidence, we include qualitative examples in Table \ref{tab:deco_comp} showing DecoRate and CLIP's performance on five example text descriptions. The results demonstrate how DecoRate's use of a multimodal LLM provides enhanced language understanding, resulting in better sensitivity to details in the provided descriptions. In the final example text prompt, we omit the word "chair" to demonstrate that without an explicit indicator of the object type, CLIP is unable to use the provided details to infer object appearance, while DecoRate still performs effectively.

\subsection{Qualitative Comparison with LayoutGPT}

As mentioned previously, LayoutGPT does not incorporate text description information into its furniture retrieval process, only selecting based on category and size. As a result, it is not designed to satisfy visual specifications. Here, we demonstrate the effect of this by visualizing renderings of scenes generated by \textsc{Decorum} and LayoutGPT based on the same text prompt.

\begin{figure}[htbp]
    \centering
    \includegraphics[width=0.5\textwidth]{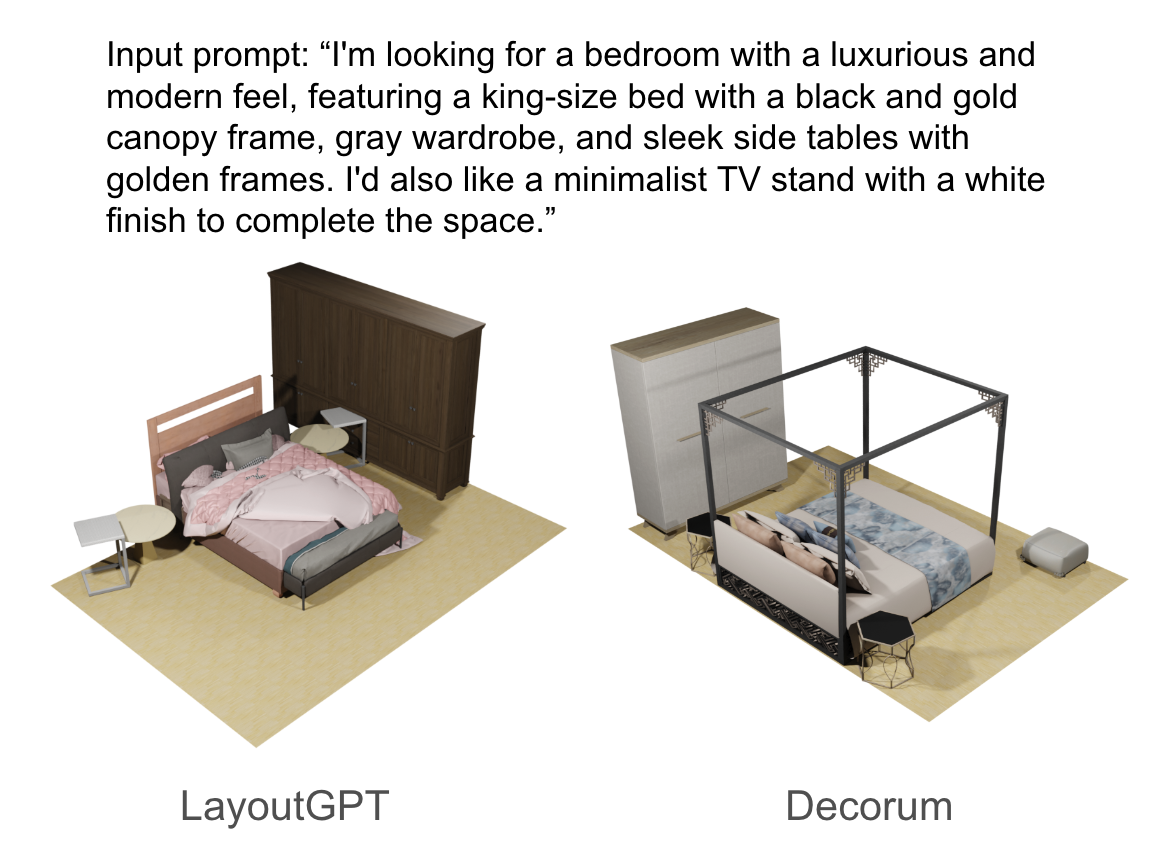}
    % \vspace{0.5cm} % Adjust the space between the images as needed
    % \includegraphics[width=\textwidth]{Figures/lr_qualitative.png}
    \caption{Example of LayoutGPT text-conditioned output compared to \textsc{Decorum} for a sample text prompt. Because LayoutGPT does not incorporate information from the text prompt into its choice of objects, it cannot satisfy visual descriptions}
    \label{fig:layoutgpt_comparison}
\end{figure}

\subsection{CLIP-Score Evaluations}

\begin{table}[h!]
\centering
\begin{tabular}{|c|c|}
\hline
\textbf{Method} & \textbf{CLIP-Score} \\ \hline
Ground-Truth & 0.258 \\ \hline
\textbf{Decorum(ours)} & \textbf{0.257} \\ \hline
CLIP-Layout & 0.255 \\ \hline
Random & 0.255 \\ \hline
\end{tabular}
\caption{CLIP-Score comparison for text-to-scene methods }
\label{tab:clipscore}
\end{table}

In this section we include CLIP-Score evaluations for our model compared with several baselines, displayed in Table \ref{tab:clipscore}. For comparison, we include the CLIP-Score for ground-truth 3D-FRONT scenes paired with their 3D-GRAND annotations as well as with randomly selected 3D-GRAND captions. We again find that our model shows greater responsiveness to text inputs compared to CLIP-Layout, which is unable to model scene-level style. However, we note that CLIP-Score is a far less informative metric compared to our introduced Text Fidelity Rating, showing little difference overall between ground-truth pairs and random pairs.

\subsection{Decorum Output Examples}
\label{sec:qual_examples}
\begin{figure*}[t!]
    \centering
    \includegraphics[width=\textwidth]{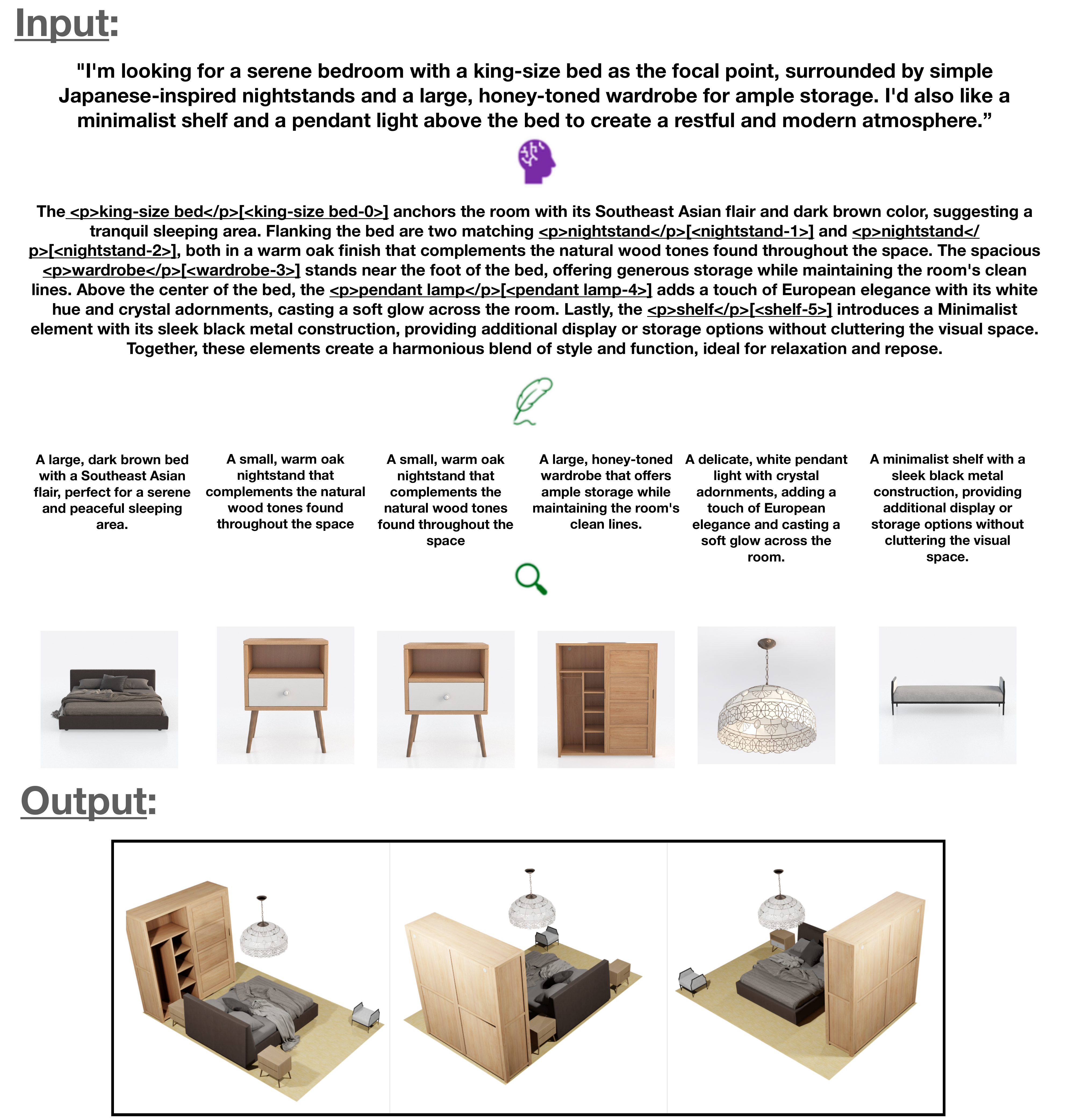}
    % \vspace{0.5cm} % Adjust the space between the images as needed
    % \includegraphics[width=\textwidth]{Figures/lr_qualitative.png}
    \caption{Example of \textsc{Decorum} pipeline applied to a sample input. For this sample prompt we show the model's predicted annotation which is used for layout generation and object selection. We then show the description generated for each object and the corresponding 3D object retrieved for this description. Finally, we include renderings of the final scene created from arranging the selected furniture.}
    \label{fig:detailed_output}
\end{figure*}

\begin{figure*}[t!]
    \centering
    \includegraphics[width=\textwidth]{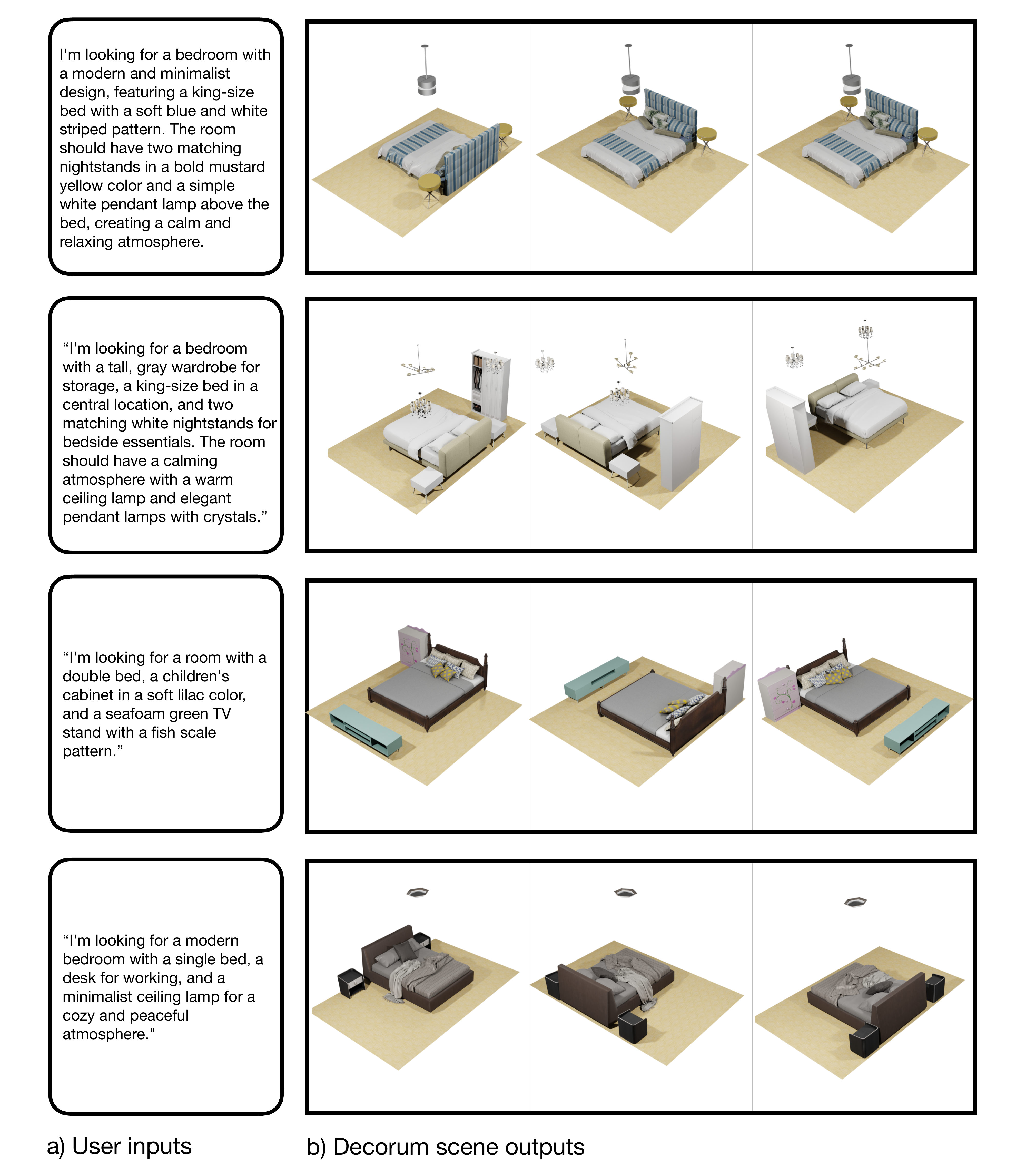}
    % \vspace{0.5cm} % Adjust the space between the images as needed
    % \includegraphics[width=\textwidth]{Figures/lr_qualitative.png}
    \caption{Visualization of scenes synthesized by Decorum to match provided user prompts}
    \label{fig:example_1}
\end{figure*}

\begin{figure*}[t!]
    \centering
    \includegraphics[width=\textwidth]{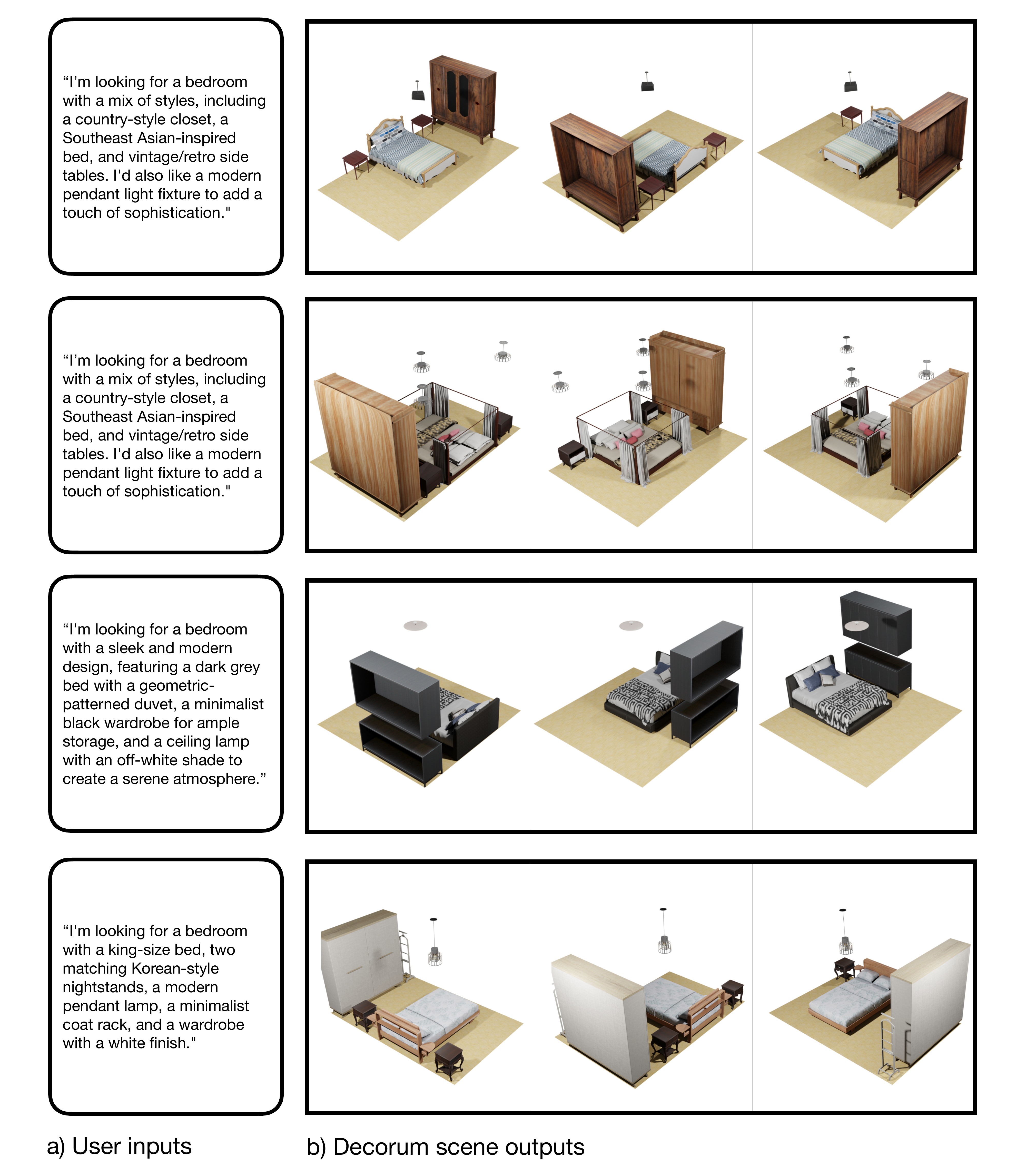}
    % \vspace{0.5cm} % Adjust the space between the images as needed
    % \includegraphics[width=\textwidth]{Figures/lr_qualitative.png}
    \caption{Visualization of scenes synthesized by Decorum to match provided user prompts}
    \label{fig:example_2}
\end{figure*}

\begin{figure*}[t!]
    \centering
    \includegraphics[width=\textwidth]{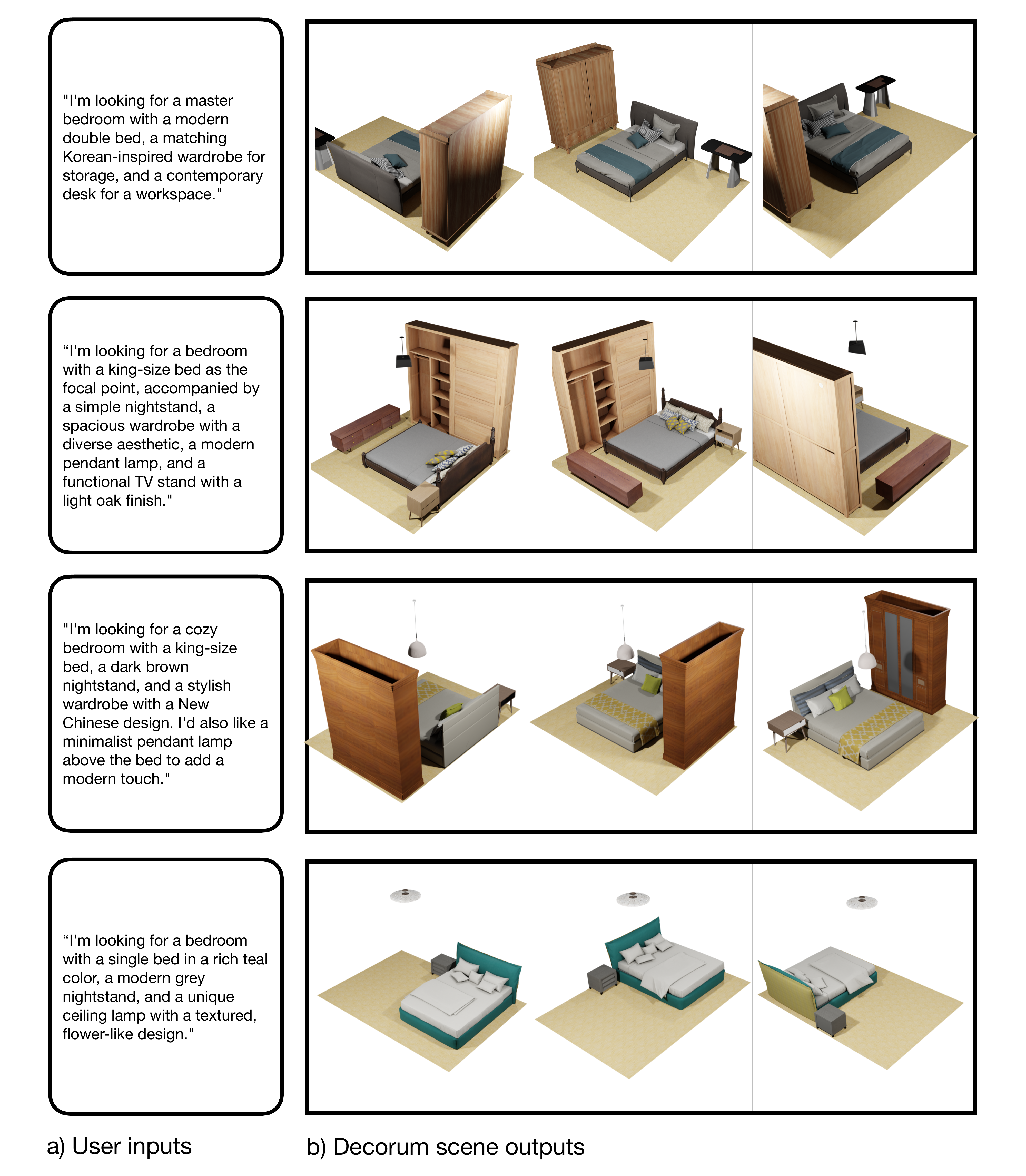}
    % \vspace{0.5cm} % Adjust the space between the images as needed
    % \includegraphics[width=\textwidth]{Figures/lr_qualitative.png}
    \caption{Visualization of scenes synthesized by Decorum to match provided user prompts}
    \label{fig:example_3}
\end{figure*}

We present additional visual examples of Decorum's generative abilities conditioned on a diverse array of user-specified styles in Fig \ref{fig:example_1},\ref{fig:example_2},\ref{fig:example_3}. To further display the intermediate steps of our generation pipeline displayed in \ref{fig:pipeline}, we include a detailed visualization of each step of the generation process for an example input in Fig \ref{fig:detailed_output}.

\input{Tables/decorate_comparison}

%% file: Tables/decorate_comparison.tex
\begin{table*}[t]
\begin{center}
\begin{adjustbox}{width=\textwidth}
\begin{tabular}{|p{0.4\textwidth}|c|c|c|c|c|c|}
\hline
\Large \textbf{Example Prompt} & \Large \textbf{Method} & \Large \textbf{Rank 1} & \Large \textbf{Rank 2} & \Large \textbf{Rank 3} & \Large \textbf{Rank 4} & \Large \textbf{Rank 5} \\ \hline

%1.
\multirow{2}{=}{\raggedright \Large \textbf{1.} The bed has a rounded blue headboard with fluffy white bedding on top. The pillows are a tasteful complimentary color to provide a cohesive and inviting feel}
& \rule[10mm]{0mm}{5mm} \Large DecoRate 
& \raisebox{-0.45\height}{\hspace{2pt}\vspace{2pt}\includegraphics[width=0.15\textwidth]{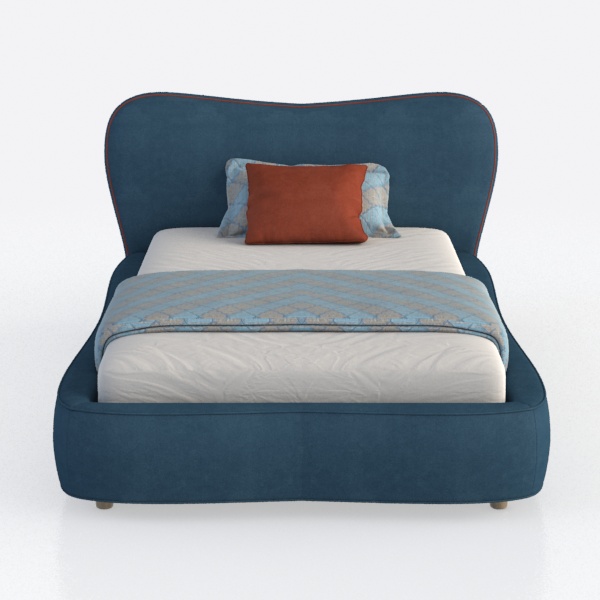}\hspace{2pt}\vspace{2pt}}
& \raisebox{-0.45\height}{\hspace{2pt}\vspace{2pt}\includegraphics[width=0.15\textwidth]{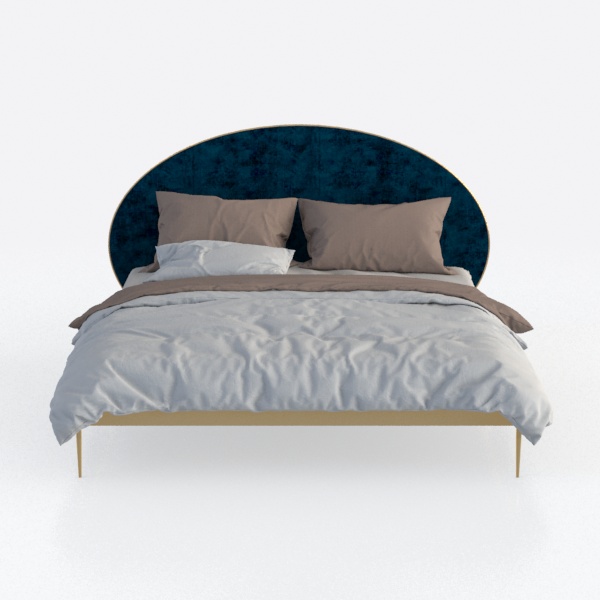}\hspace{2pt}\vspace{2pt}}
& \raisebox{-0.45\height}{\hspace{2pt}\vspace{2pt}\includegraphics[width=0.15\textwidth]{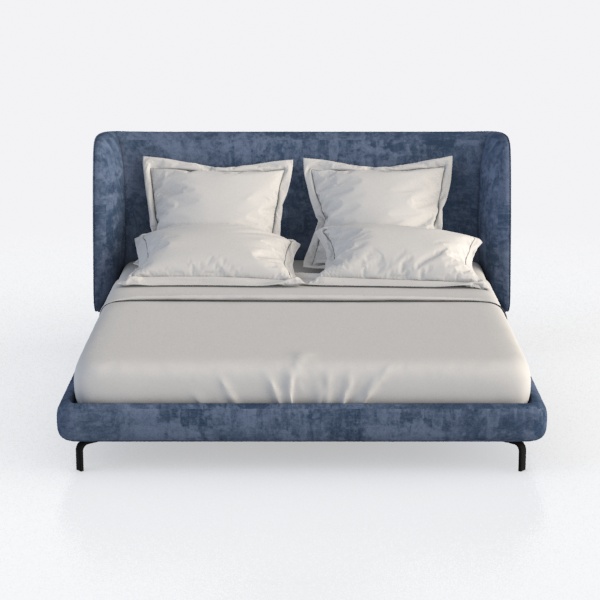}\hspace{2pt}\vspace{2pt}}
& \raisebox{-0.45\height}{\hspace{2pt}\vspace{2pt}\includegraphics[width=0.15\textwidth]{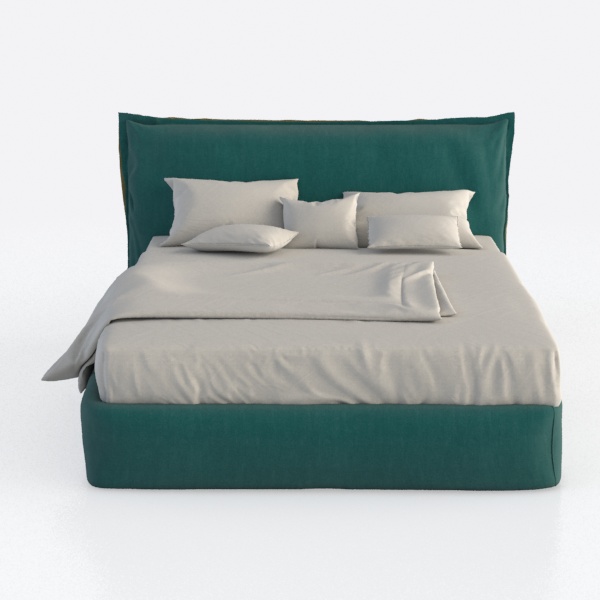}\hspace{2pt}\vspace{2pt}}
& \raisebox{-0.45\height}{\hspace{2pt}\vspace{2pt}\includegraphics[width=0.15\textwidth]{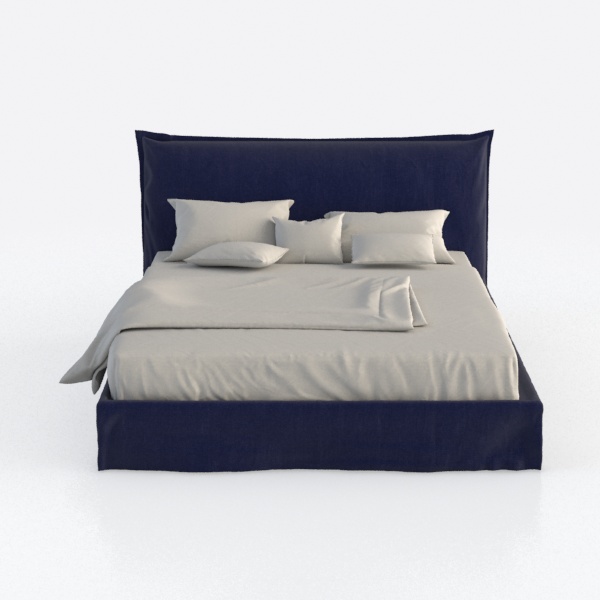}\hspace{2pt}\vspace{2pt}} \\ \cline{2-7}
& \rule[10mm]{0mm}{5mm} \Large CLIP
& \raisebox{-0.45\height}{\hspace{2pt}\vspace{2pt}\includegraphics[width=0.15\textwidth]{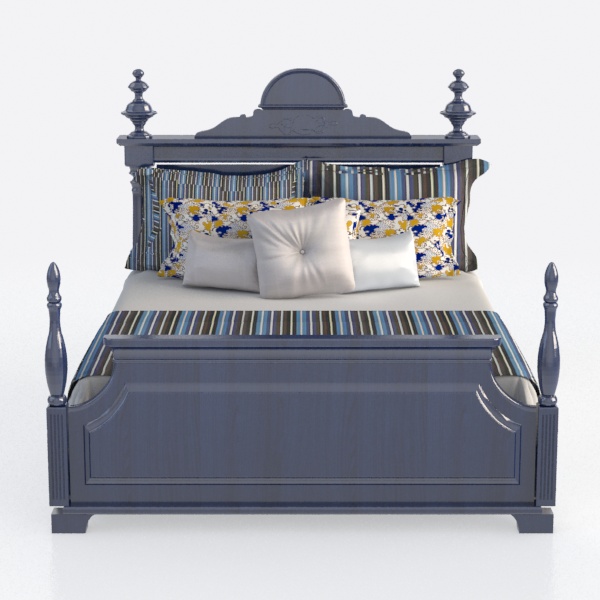}\hspace{2pt}\vspace{2pt}}
& \raisebox{-0.45\height}{\hspace{2pt}\vspace{2pt}\includegraphics[width=0.15\textwidth]{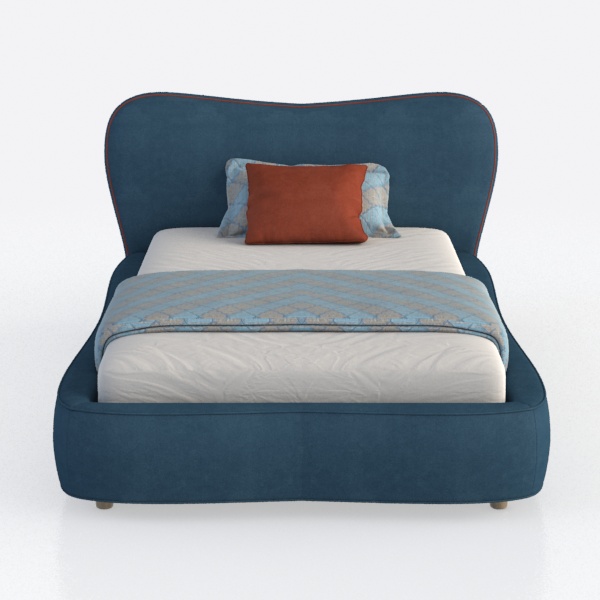}\hspace{2pt}\vspace{2pt}}
& \raisebox{-0.45\height}{\hspace{2pt}\vspace{2pt}\includegraphics[width=0.15\textwidth]{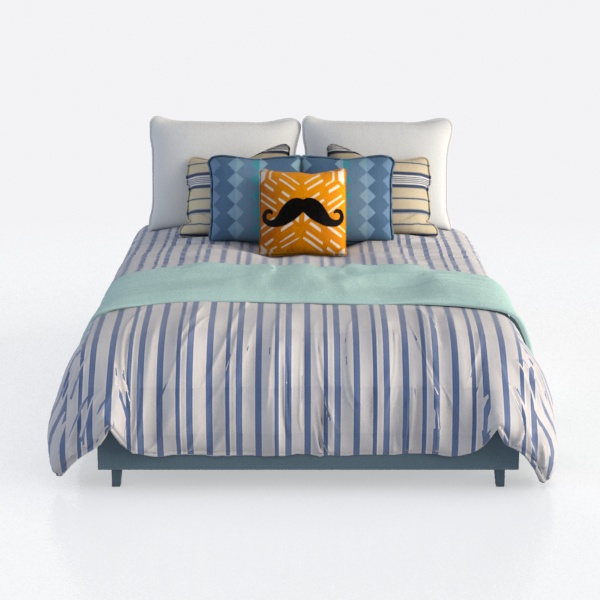}\hspace{2pt}\vspace{2pt}}
& \raisebox{-0.45\height}{\hspace{2pt}\vspace{2pt}\includegraphics[width=0.15\textwidth]{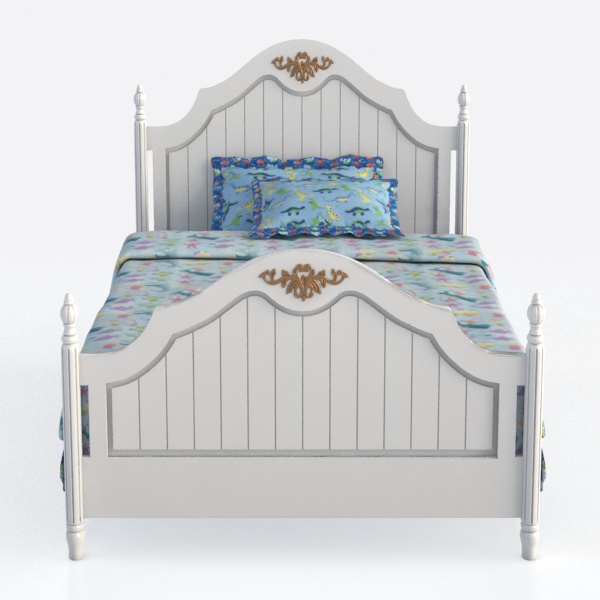}\hspace{2pt}\vspace{2pt}}
& \raisebox{-0.45\height}{\hspace{2pt}\vspace{2pt}\includegraphics[width=0.15\textwidth]{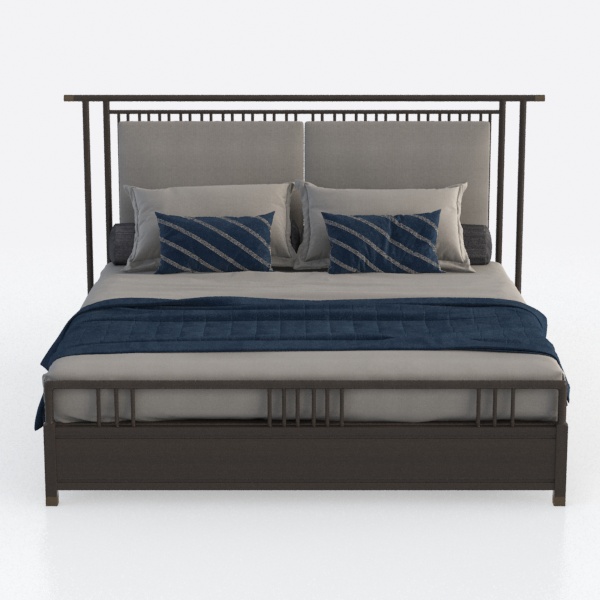}\hspace{2pt}\vspace{2pt}} \\ \hline

%2. 
\multirow{2}{=}{\Large \textbf{2.} The grey couch is stacked with plenty of pillows and features two sections}
& \rule[10mm]{0mm}{5mm} \Large DecoRate 
& \raisebox{-0.45\height}{\hspace{2pt}\vspace{2pt}\includegraphics[width=0.15\textwidth]{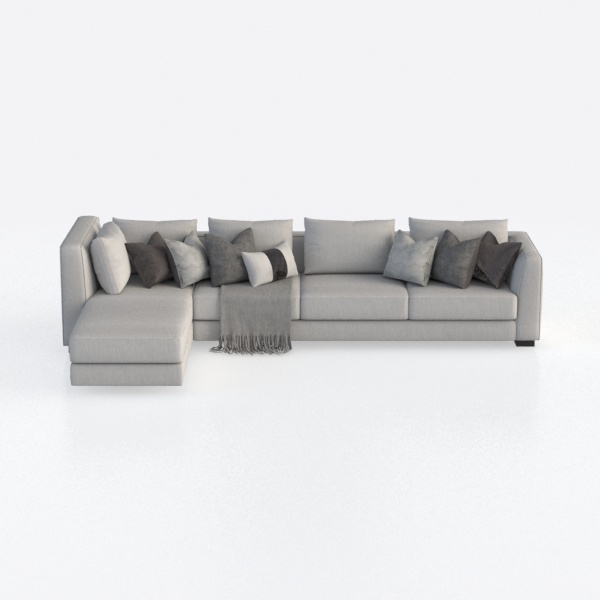}\hspace{2pt}\vspace{2pt}}
& \raisebox{-0.45\height}{\hspace{2pt}\vspace{2pt}\includegraphics[width=0.15\textwidth]{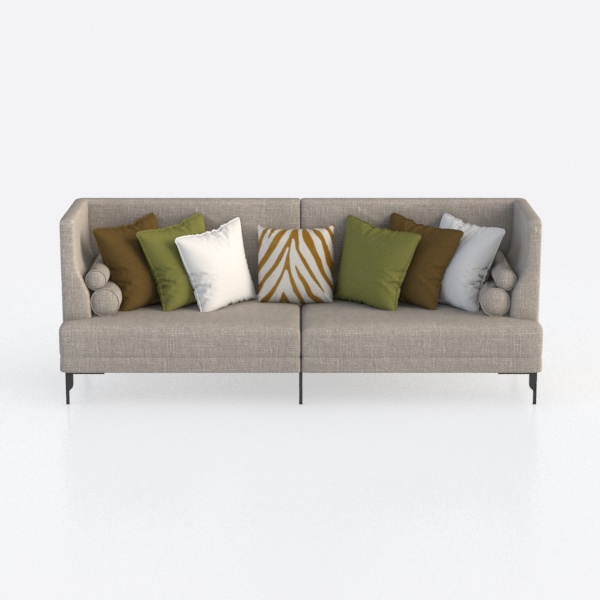}\hspace{2pt}\vspace{2pt}}
& \raisebox{-0.45\height}{\hspace{2pt}\vspace{2pt}\includegraphics[width=0.15\textwidth]{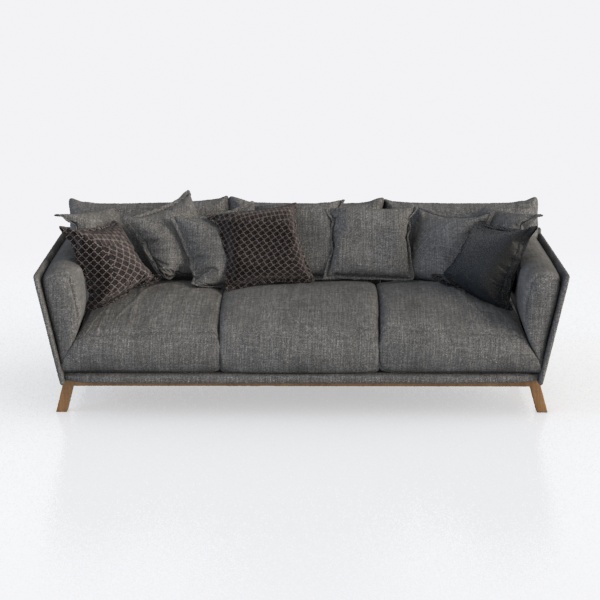}\hspace{2pt}\vspace{2pt}}
& \raisebox{-0.45\height}{\hspace{2pt}\vspace{2pt}\includegraphics[width=0.15\textwidth]{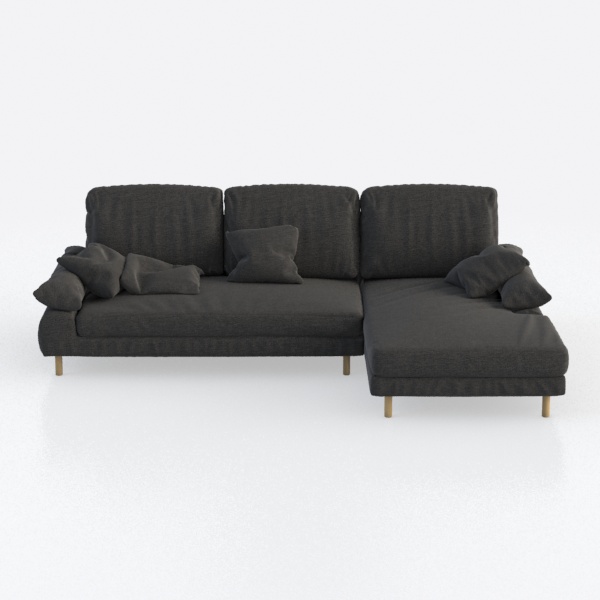}\hspace{2pt}\vspace{2pt}}
& \raisebox{-0.45\height}{\hspace{2pt}\vspace{2pt}\includegraphics[width=0.15\textwidth]{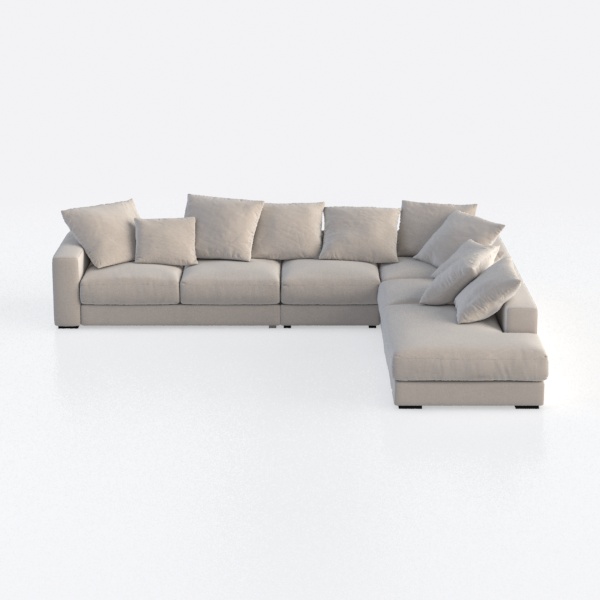}\hspace{2pt}\vspace{2pt}} \\ \cline{2-7}
& \rule[10mm]{0mm}{5mm} \Large CLIP
& \raisebox{-0.45\height}{\hspace{2pt}\vspace{2pt}\includegraphics[width=0.15\textwidth]{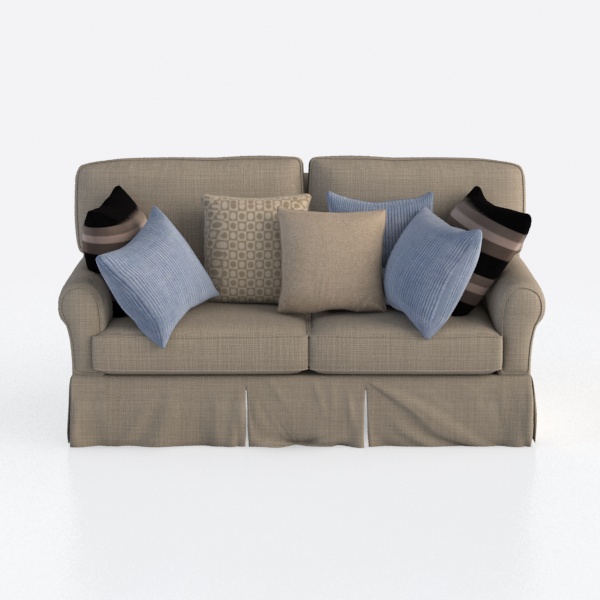}\hspace{2pt}\vspace{2pt}}
& \raisebox{-0.45\height}{\hspace{2pt}\vspace{2pt}\includegraphics[width=0.15\textwidth]{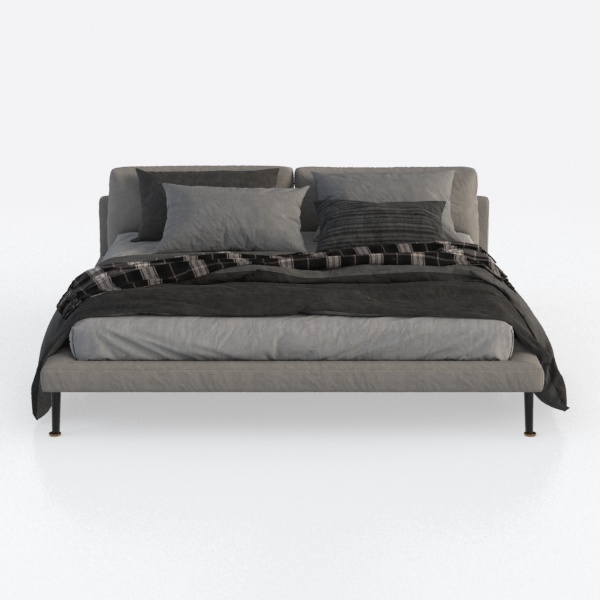}\hspace{2pt}\vspace{2pt}}
& \raisebox{-0.45\height}{\hspace{2pt}\vspace{2pt}\includegraphics[width=0.15\textwidth]{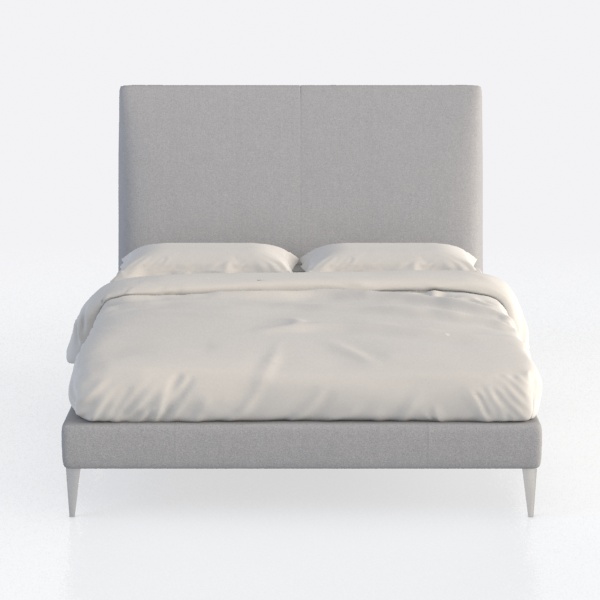}\hspace{2pt}\vspace{2pt}}
& \raisebox{-0.45\height}{\hspace{2pt}\vspace{2pt}\includegraphics[width=0.15\textwidth]{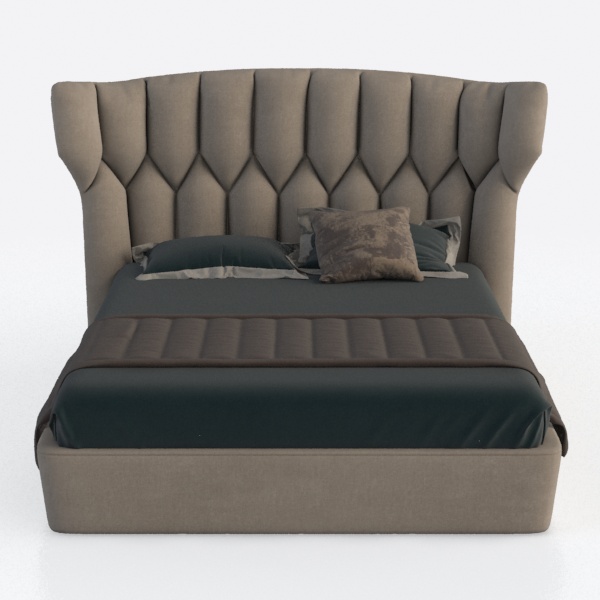}\hspace{2pt}\vspace{2pt}}
& \raisebox{-0.45\height}{\hspace{2pt}\vspace{2pt}\includegraphics[width=0.15\textwidth]{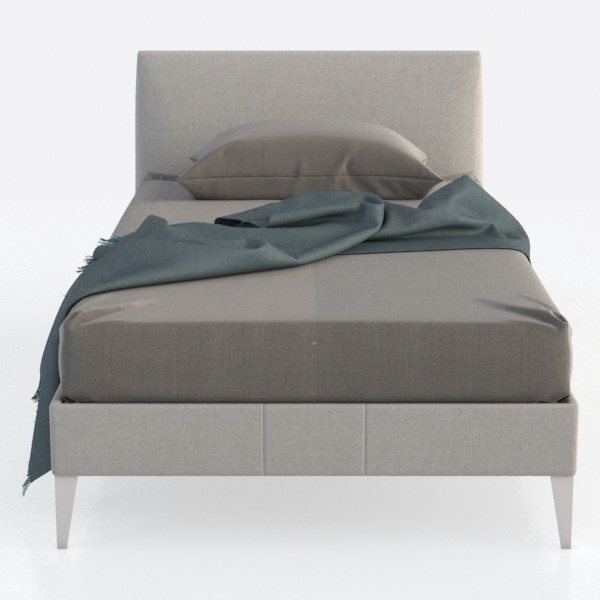}\hspace{2pt}\vspace{2pt}} \\ \hline

%3. 
\multirow{2}{=}{\Large \textbf{3.} The object is a small potted plants with bushy green leaves shooting out at the top}
& \rule[10mm]{0mm}{5mm} \Large DecoRate 
& \raisebox{-0.45\height}{\hspace{2pt}\vspace{2pt}\includegraphics[width=0.15\textwidth]{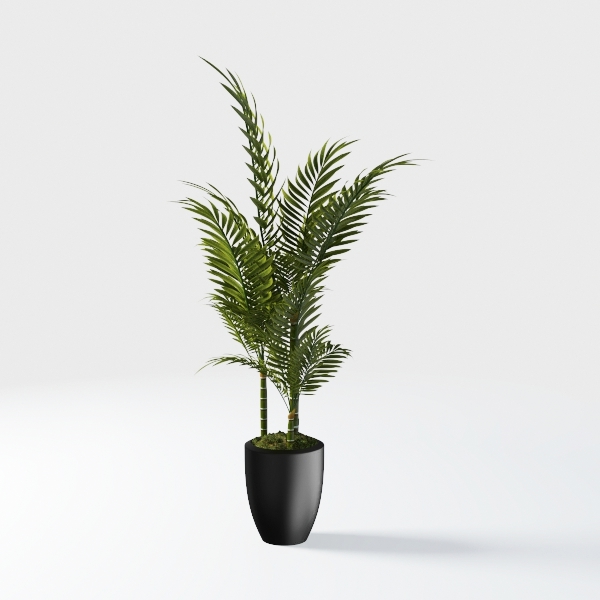}\hspace{2pt}\vspace{2pt}}
& \raisebox{-0.45\height}{\hspace{2pt}\vspace{2pt}\includegraphics[width=0.15\textwidth]{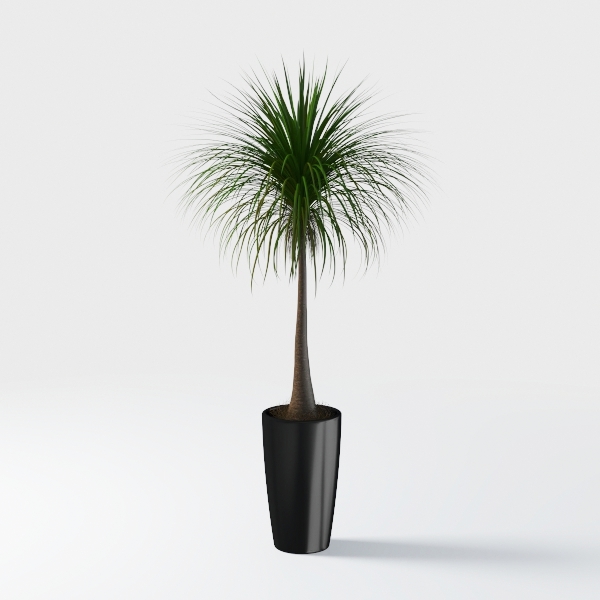}\hspace{2pt}\vspace{2pt}}
& \raisebox{-0.45\height}{\hspace{2pt}\vspace{2pt}\includegraphics[width=0.15\textwidth]{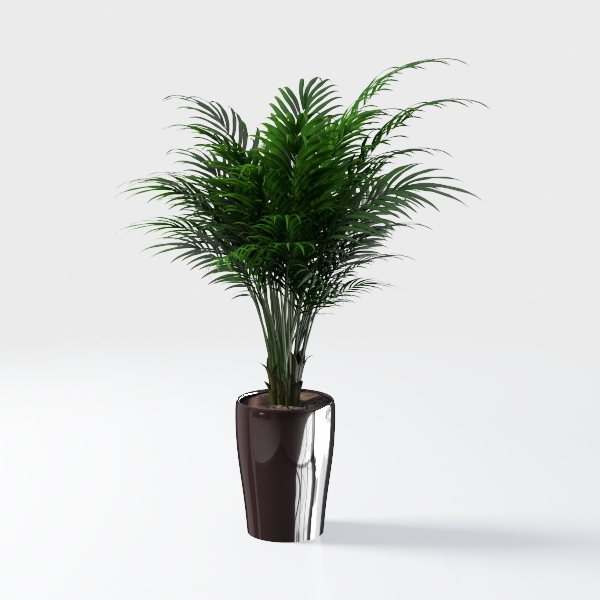}\hspace{2pt}\vspace{2pt}}
& \raisebox{-0.45\height}{\hspace{2pt}\vspace{2pt}\includegraphics[width=0.15\textwidth]{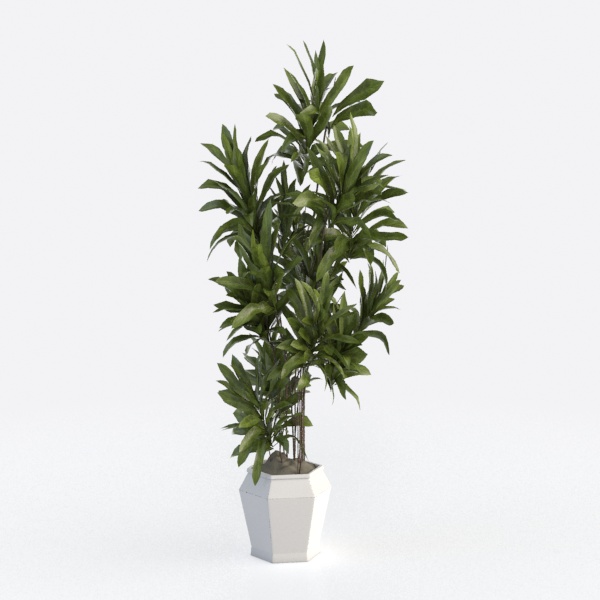}\hspace{2pt}\vspace{2pt}}
& \raisebox{-0.45\height}{\hspace{2pt}\vspace{2pt}\includegraphics[width=0.15\textwidth]{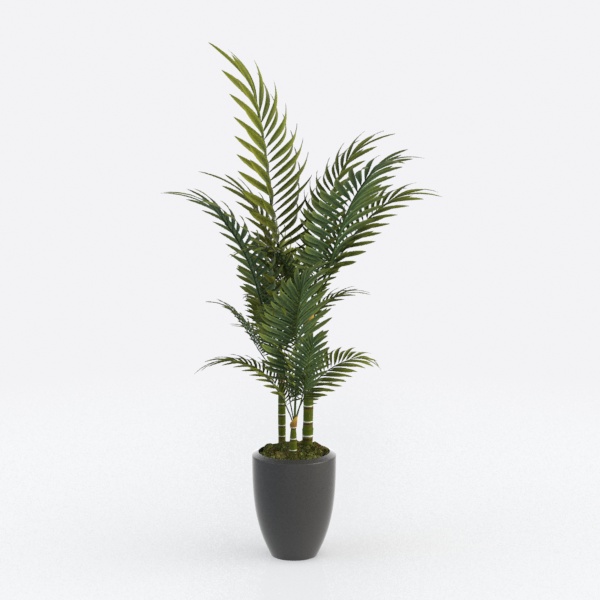}\hspace{2pt}\vspace{2pt}} \\ \cline{2-7}
& \rule[10mm]{0mm}{5mm} \Large CLIP
& \raisebox{-0.45\height}{\hspace{2pt}\vspace{2pt}\includegraphics[width=0.15\textwidth]{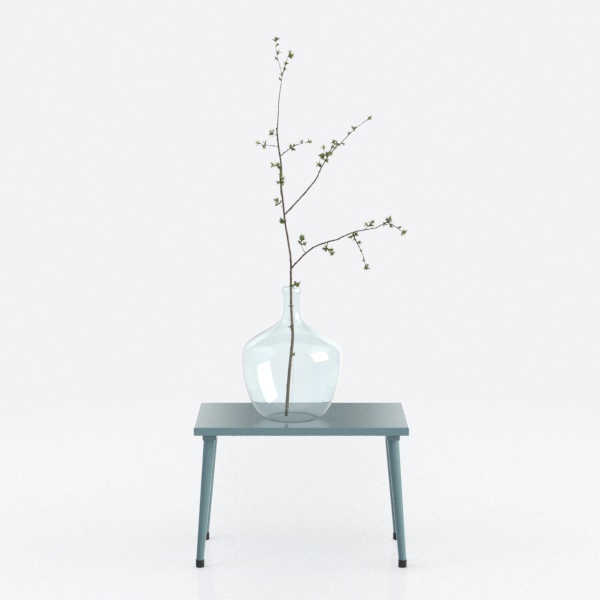}\hspace{2pt}\vspace{2pt}}
& \raisebox{-0.45\height}{\hspace{2pt}\vspace{2pt}\includegraphics[width=0.15\textwidth]{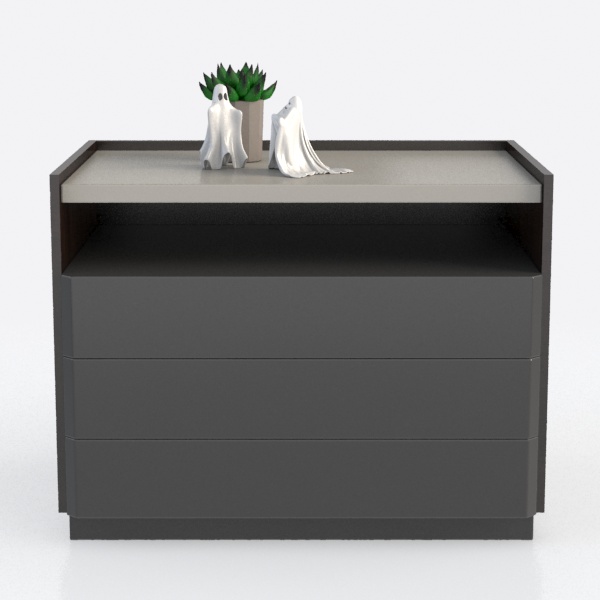}\hspace{2pt}\vspace{2pt}}
& \raisebox{-0.45\height}{\hspace{2pt}\vspace{2pt}\includegraphics[width=0.15\textwidth]{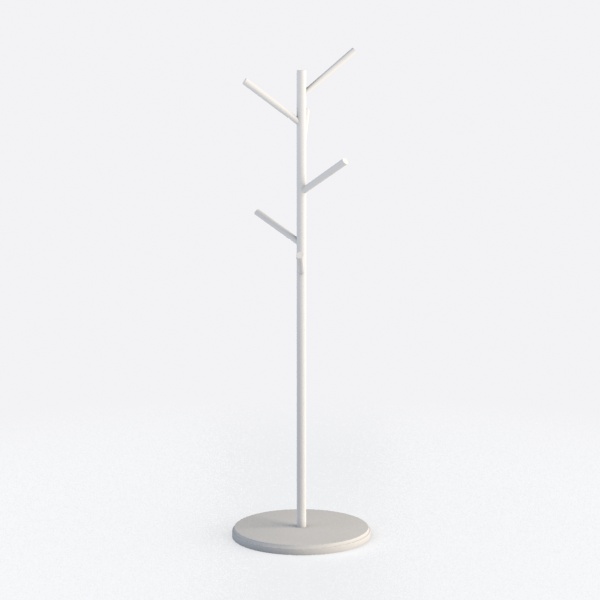}\hspace{2pt}\vspace{2pt}}
& \raisebox{-0.45\height}{\hspace{2pt}\vspace{2pt}\includegraphics[width=0.15\textwidth]{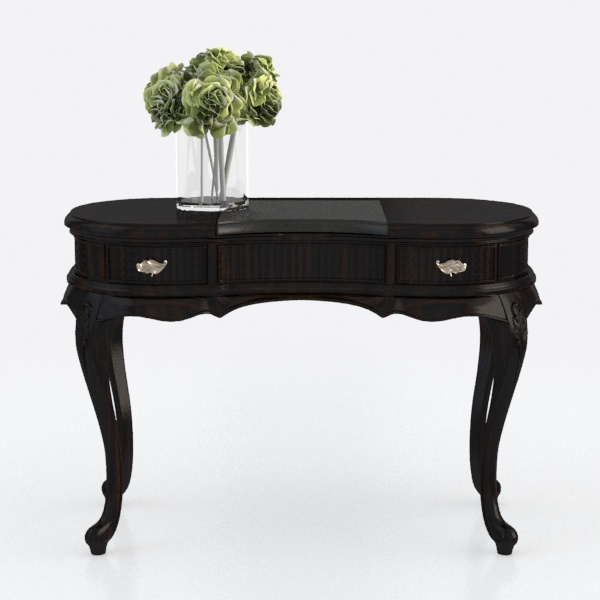}\hspace{2pt}\vspace{2pt}}
& \raisebox{-0.45\height}{\hspace{2pt}\vspace{2pt}\includegraphics[width=0.15\textwidth]{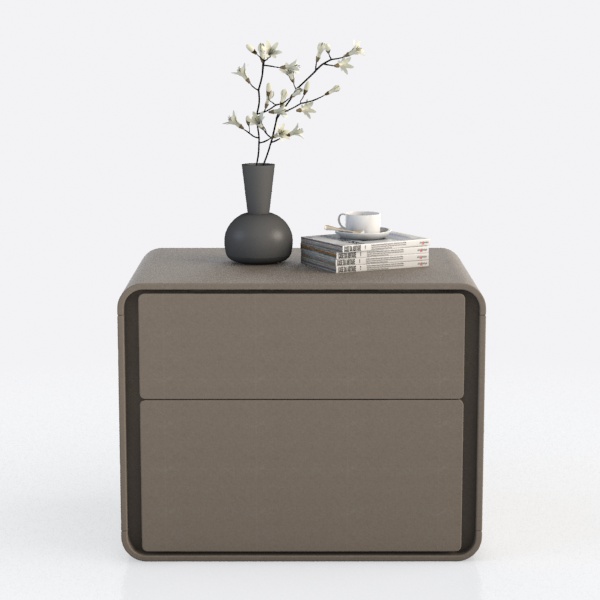}\hspace{2pt}\vspace{2pt}} \\ \hline

%4. 
\multirow{2}{=} {\Large \textbf{4.} The chandelier is a truly dazzling piece with a gold frame holding lights evenly}
& \rule[10mm]{0mm}{5mm} \Large DecoRate 
& \raisebox{-0.45\height}{\hspace{2pt}\vspace{2pt}\includegraphics[width=0.15\textwidth]{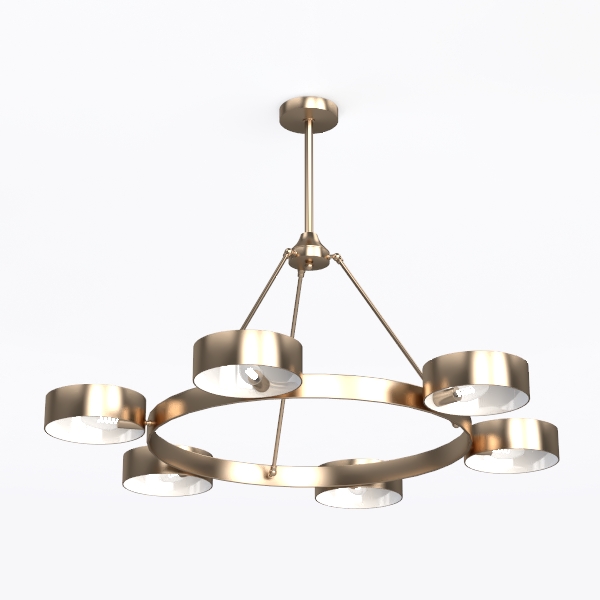}\hspace{2pt}\vspace{2pt}}
& \raisebox{-0.45\height}{\hspace{2pt}\vspace{2pt}\includegraphics[width=0.15\textwidth]{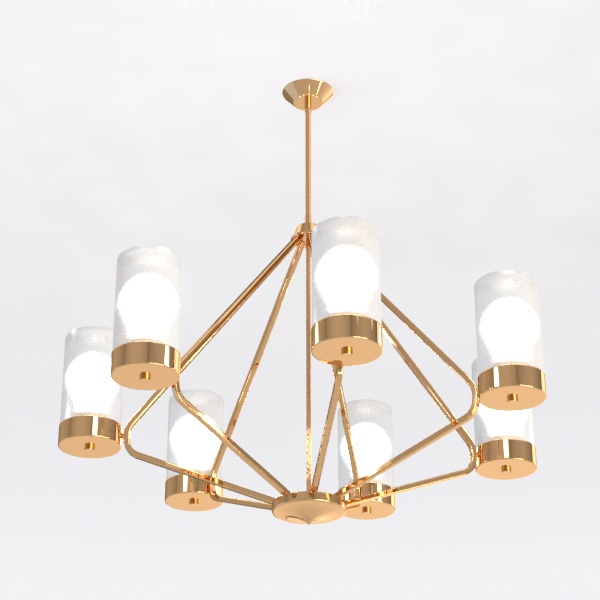}\hspace{2pt}\vspace{2pt}}
& \raisebox{-0.45\height}{\hspace{2pt}\vspace{2pt}\includegraphics[width=0.15\textwidth]{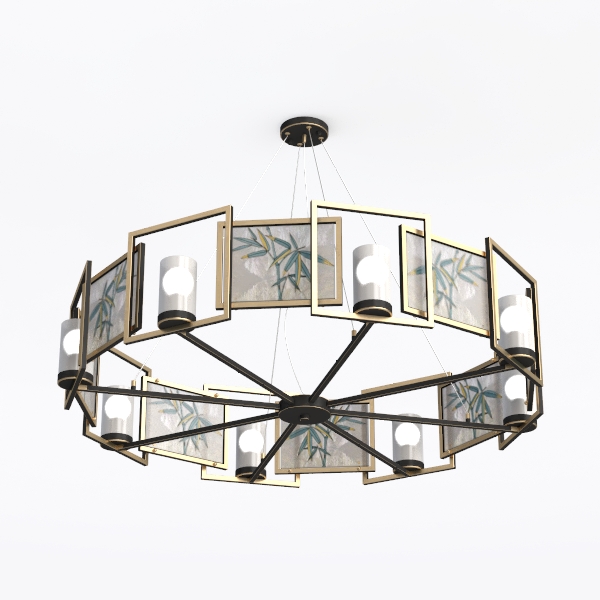}\hspace{2pt}\vspace{2pt}}
& \raisebox{-0.45\height}{\hspace{2pt}\vspace{2pt}\includegraphics[width=0.15\textwidth]{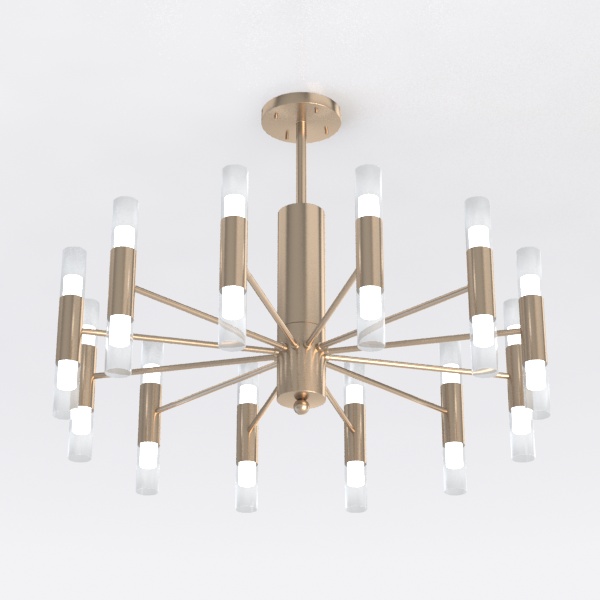}\hspace{2pt}\vspace{2pt}}
& \raisebox{-0.45\height}{\hspace{2pt}\vspace{2pt}\includegraphics[width=0.15\textwidth]{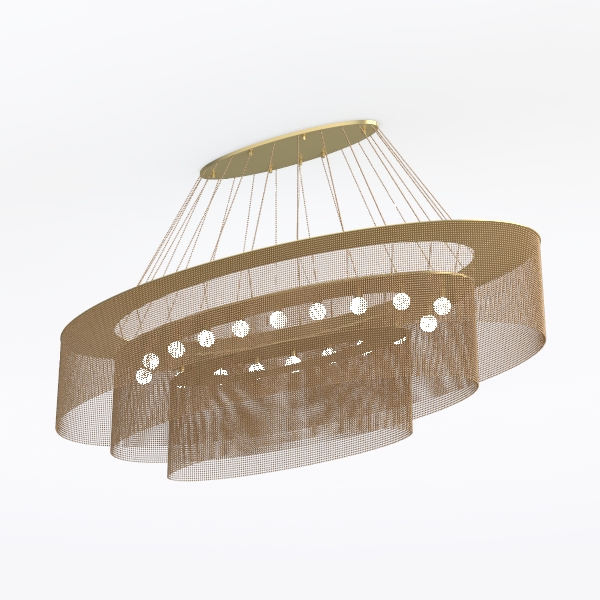}\hspace{2pt}\vspace{2pt}} \\ \cline{2-7}
& \rule[10mm]{0mm}{5mm} \Large CLIP
& \raisebox{-0.45\height}{\hspace{2pt}\vspace{2pt}\includegraphics[width=0.15\textwidth]{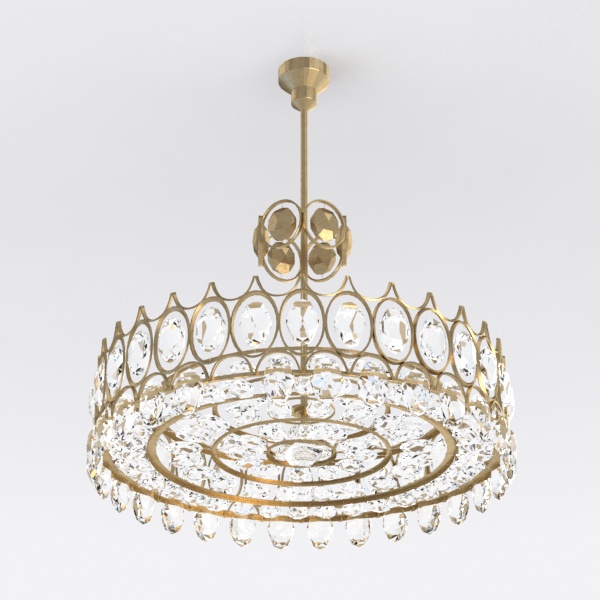}\hspace{2pt}\vspace{2pt}}
& \raisebox{-0.45\height}{\hspace{2pt}\vspace{2pt}\includegraphics[width=0.15\textwidth]{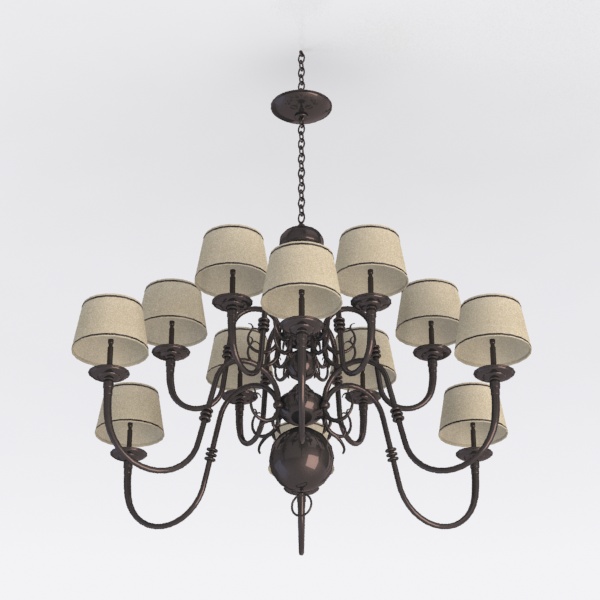}\hspace{2pt}\vspace{2pt}}
& \raisebox{-0.45\height}{\hspace{2pt}\vspace{2pt}\includegraphics[width=0.15\textwidth]{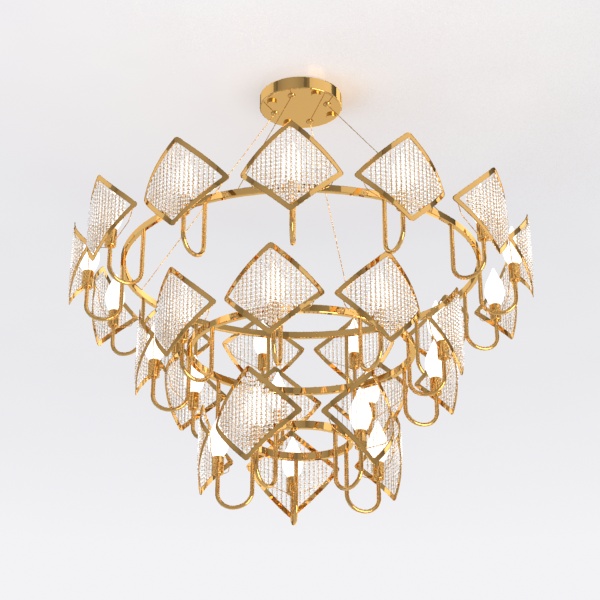}\hspace{2pt}\vspace{2pt}}
& \raisebox{-0.45\height}{\hspace{2pt}\vspace{2pt}\includegraphics[width=0.15\textwidth]{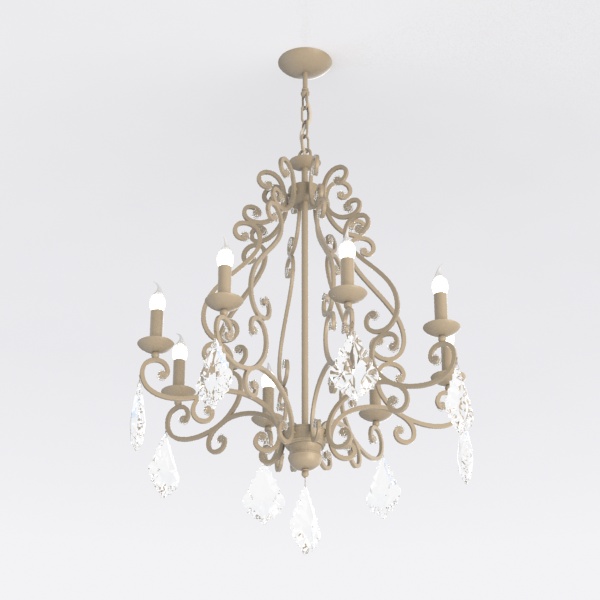}\hspace{2pt}\vspace{2pt}}
& \raisebox{-0.45\height}{\hspace{2pt}\vspace{2pt}\includegraphics[width=0.15\textwidth]{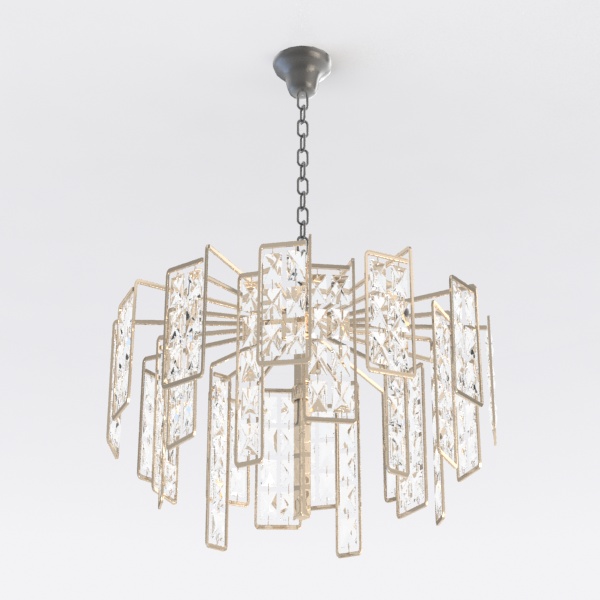}\hspace{2pt}\vspace{2pt}} \\ \hline

%5. 
\multirow{2}{=}{\Large \textbf{5.} The object is a dark blue office with wheels on the bottom that allow it to move around }
& \rule[10mm]{0mm}{5mm} \Large DecoRate 
& \raisebox{-0.45\height}{\hspace{2pt}\vspace{2pt}\includegraphics[width=0.15\textwidth]{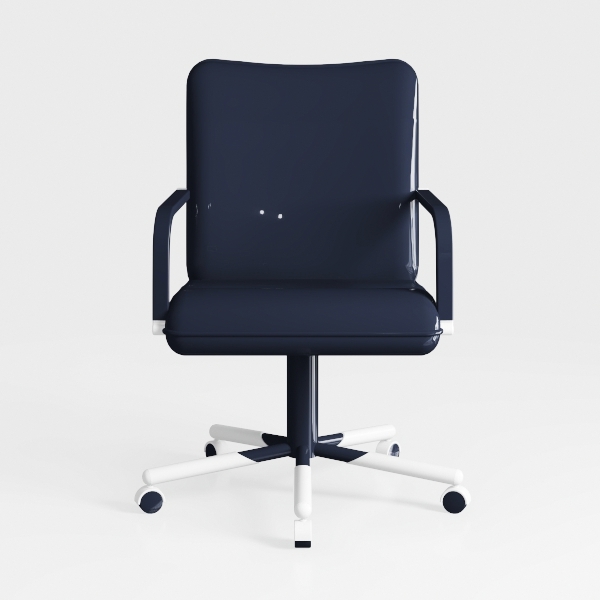}\hspace{2pt}\vspace{2pt}}
& \raisebox{-0.45\height}{\hspace{2pt}\vspace{2pt}\includegraphics[width=0.15\textwidth]{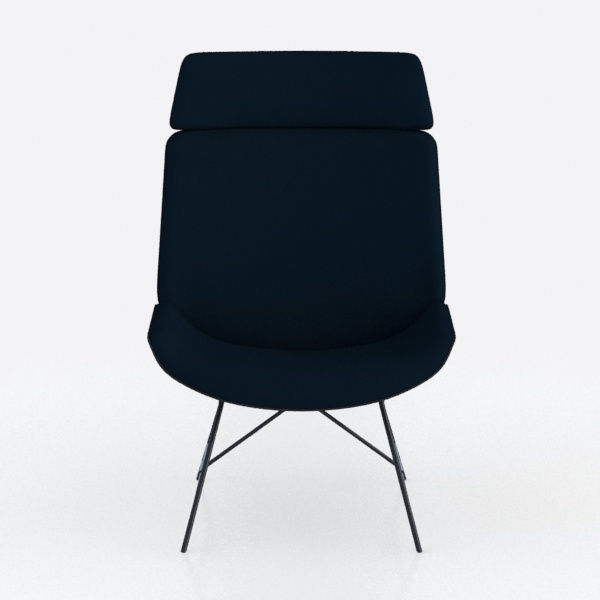}\hspace{2pt}\vspace{2pt}}
& \raisebox{-0.45\height}{\hspace{2pt}\vspace{2pt}\includegraphics[width=0.15\textwidth]{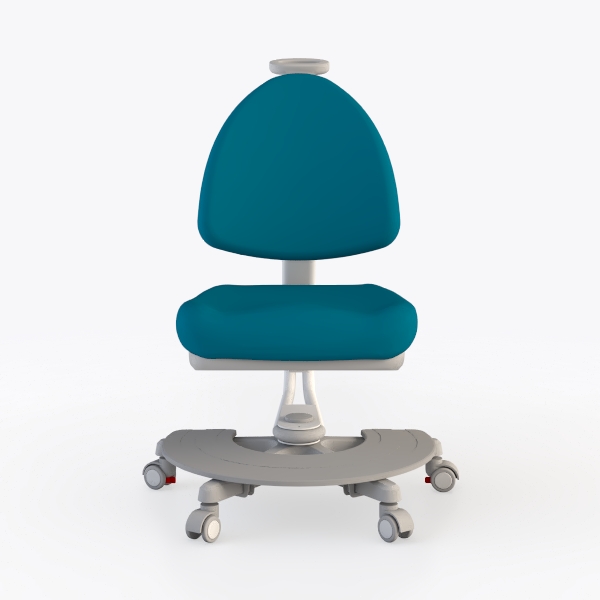}\hspace{2pt}\vspace{2pt}}
& \raisebox{-0.45\height}{\hspace{2pt}\vspace{2pt}\includegraphics[width=0.15\textwidth]{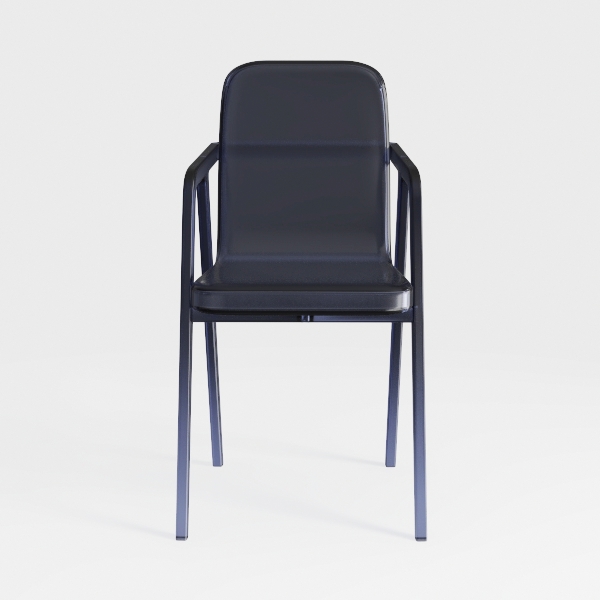}\hspace{2pt}\vspace{2pt}}
& \raisebox{-0.45\height}{\hspace{2pt}\vspace{2pt}\includegraphics[width=0.15\textwidth]{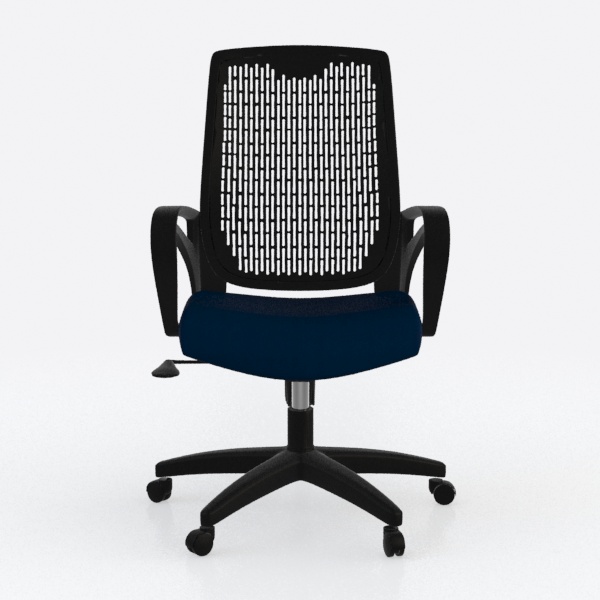}\hspace{2pt}\vspace{2pt}} \\ \cline{2-7}
& \rule[10mm]{0mm}{5mm} \Large CLIP
& \raisebox{-0.45\height}{\hspace{2pt}\vspace{2pt}\includegraphics[width=0.15\textwidth]{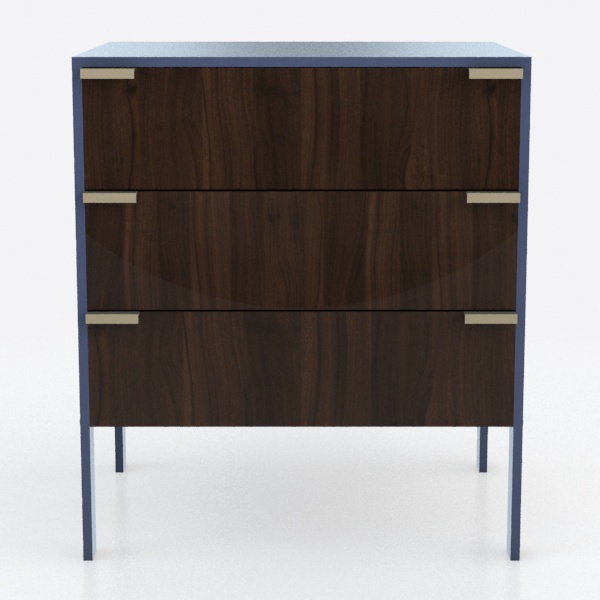}\hspace{2pt}\vspace{2pt}}
& \raisebox{-0.45\height}{\hspace{2pt}\vspace{2pt}\includegraphics[width=0.15\textwidth]{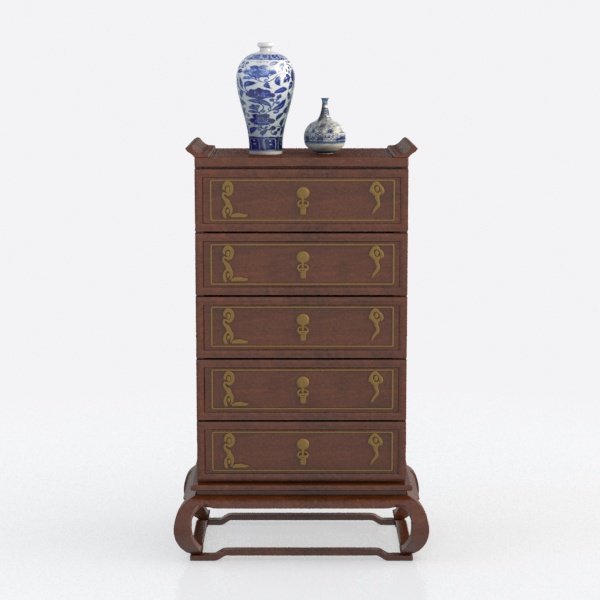}\hspace{2pt}\vspace{2pt}}
& \raisebox{-0.45\height}{\hspace{2pt}\vspace{2pt}\includegraphics[width=0.15\textwidth]{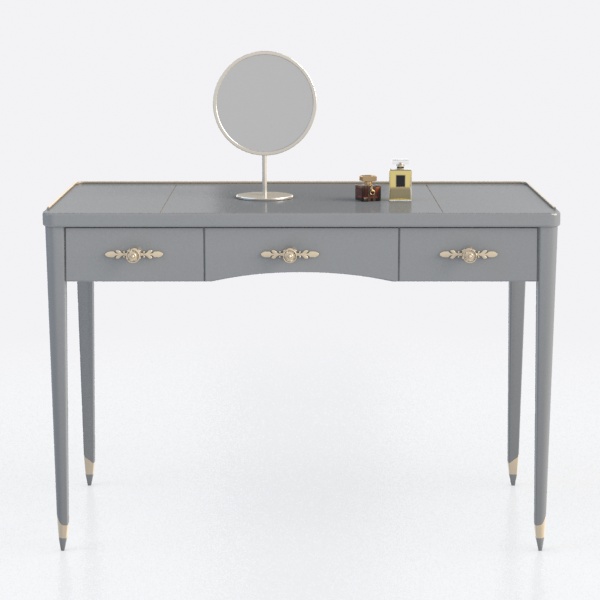}\hspace{2pt}\vspace{2pt}}
& \raisebox{-0.45\height}{\hspace{2pt}\vspace{2pt}\includegraphics[width=0.15\textwidth]{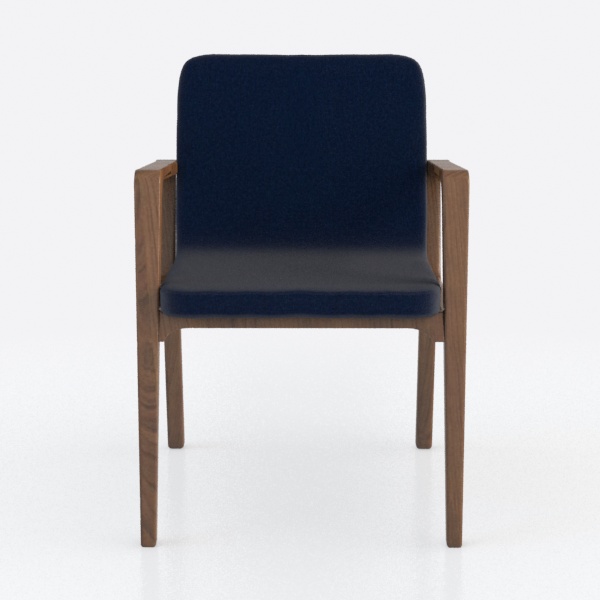}\hspace{2pt}\vspace{2pt}}
& \raisebox{-0.45\height}{\hspace{2pt}\vspace{2pt}\includegraphics[width=0.15\textwidth]{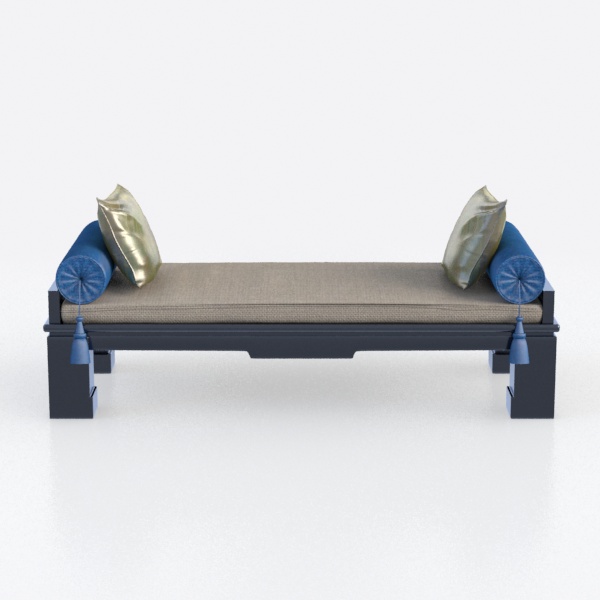}\hspace{2pt}\vspace{2pt}} \\ \hline

% Repeat the same structure for the remaining rows

\end{tabular}
\end{adjustbox}
\end{center}
\caption{Comparison of DecoRate with CLIP for object retrieval on five example text descriptions. For each object description, we show the top five objects which each model rates as having the highest similarity to the description.}
\label{tab:deco_comp}
\end{table*}